%% file: arxiv.tex
\documentclass{article}

\usepackage{microtype}
\usepackage{graphicx}
\usepackage{subfigure}
\usepackage{booktabs} 
\usepackage{listings}
\usepackage{hyperref}

\usepackage[table]{xcolor}

\usepackage[accepted]{arxiv}
\input{math_commands.tex}

\usepackage{enumitem}
\setlist[itemize]{noitemsep, left=0em}
\usepackage[utf8]{inputenc} 
\usepackage[T1]{fontenc}    
\usepackage{hyperref}       
\usepackage{url}            
\usepackage{booktabs}       
\usepackage{amsfonts}       
\usepackage{nicefrac}       
\usepackage{microtype}      
\usepackage{xcolor}         
\usepackage{microtype}
\usepackage{graphicx}
\usepackage{subfigure}
\usepackage{amsmath}
\usepackage{amssymb}
\usepackage{mathtools}
\usepackage{amsthm}
\theoremstyle{plain}

\theoremstyle{definition}

\theoremstyle{remark}

\DeclareMathOperator{\erf}{erf}
\newcommand {\myvec}[1] {{\mbox{\boldmath $#1$}}}

\usepackage{adjustbox}
\usepackage[capitalize,noabbrev]{cleveref}

\usepackage[textsize=tiny]{todonotes}

\icmltitlerunning{Interpretable Deep Clustering for Tabular Data}

\begin{document}

\twocolumn[
\icmltitle{Interpretable Deep Clustering for Tabular Data}

\begin{icmlauthorlist}
\icmlauthor{Jonathan Svirsky}{yyy}
\icmlauthor{Ofir Lindenbaum}{yyy}
\end{icmlauthorlist}

\icmlaffiliation{yyy}{Department of Engineering, Bar Ilan University, Ramat-Gan, Israel}

\icmlcorrespondingauthor{Jonathan Svirsky}{svirskj@biu.ac.il}
\icmlcorrespondingauthor{Ofir Lindenbaum}{ofir.lindenbaum@biu.ac.il}

\icmlkeywords{Deep Clustering, Interpretability}

\vskip 0.3in
]

\printAffiliationsAndNotice{} 

\begin{abstract}
Clustering is a fundamental learning task widely used as a first step in data analysis. For example, biologists use cluster assignments to analyze genome sequences, medical records, or images. Since downstream analysis is typically performed at the cluster level, practitioners seek reliable and interpretable clustering models. We propose a new deep-learning framework for general domain tabular data that predicts interpretable cluster assignments at the instance and cluster levels. First, we present a self-supervised procedure to identify the subset of the most informative features from each data point. Then, we design a model that predicts cluster assignments and a gate matrix that provides cluster-level feature selection. Overall, our model provides cluster assignments with an indication of the driving feature for each sample and each cluster. We show that the proposed method can reliably predict cluster assignments in biological, text, image, and physics tabular datasets. Furthermore, using previously proposed metrics, we verify that our model leads to interpretable results at a sample and cluster level. Our code is available on \href{https://github.com/jsvir/idc}{Github}.
\end{abstract}

\section{Introduction}
\label{sec:intro}

\paragraph{Clustering} is a crucial task in data science that helps researchers uncover and study latent structures in complex data. By grouping related data points into clusters, researchers can gain insights into the underlying characteristics of the data and identify relationships between samples and variables.
Clustering is used in various scientific fields, including biology \citep{reddy2018clustering}, physics \citep{bregman2021array}, and social sciences \citep{varghese2010clustering}. For instance, in biology, clustering can identify clones of related B cells \cite{lindenbaum2021alignment}. In psychology, based on survey data, clustering can identify different types of behavior or personality traits.

Clustering is a common technique used in bio-medicine to analyze gene expression data. It involves identifying groups of genes that have similar expression patterns across different samples. Scientists often cluster high-dimensional points corresponding to individual cells to recover known cell populations and discover new, potentially rare cell types. However, bio-med gene expression data is generally represented in a tabular, high-dimensional format, making it difficult to obtain accurate clusters with meaningful structures. In addition, interpretability is a crucial requirement for real-world bio-med datasets since it is essential to understand the biological meaning behind the identified clusters. As a result, there is an increasing demand in bio-medicine for clustering models that offer interpretability for tabular data.

In bio-medicine, clustering is widely used to analyze gene expression data, where cluster assignments can identify groups of genes with similar expression patterns across different samples \citep{armingol2021deciphering}. When applied to single-cell omics data \citep{wang2010single}, clustering recovers known cell populations while discovering new and perhaps rare cell types \citep{deprez2020single}. Unfortunately, such bio-med gene expression data types are generally represented in a tabular, high-dimensional format, making it difficult to obtain accurate clusters with meaningful structures. In addition, interpretability is a crucial requirement for real-world bio-med datasets since it is essential to understand the biological meaning behind the identified clusters \citep{yang2021feature}. As a result, there is an increasing demand in bio-medicine for clustering models that offer interpretability for tabular data.

\paragraph{Interpretability} in machine learning refers to the ability to understand and explain the predictions and decisions made by predictive models. It is critical for the proper deployment of machine learning systems, especially in applications where transparency and accountability are essential. Interpretability can take different forms, including interpretable model structure, identification of feature importance for model predictions, visualization of data, and generation of explanations for the prediction. In this work, we aim to design a model that achieves interpretability by sample-wise feature selection and generating cluster-level interpretations of model results. This type of interpretability is crucial for biomedical applications, for example, when seeking marker genes that are "typical" for different clusters in high-dimensional biological measurements.

In recent years, the use of deep learning models for clustering has been gaining interest \citep{shen2021you, li2022neural, cai2022efficient, niu2021spice}. These models offer better clustering capabilities by providing an improved embedding of data points. However, most existing schemes focus on image data, require domain-specific augmentations, and are not interpretable. Interpretability has also been gaining attention in deep learning, but most models focus on supervised learning \cite{alvarez2018towards,yoon2019invase,yang2022locally}.  We aim to extend these ideas to unsupervised learning by developing a deep clustering model that is interpretable by design and can be applied to general domain data. In this context, interpretability means being able to identify variables that \textit{drive} the formation of clusters in the data \citep{bertsimas2021interpretable}.  We demonstrate the effectiveness of our method in terms of clustering accuracy and interpretability in biomed, physics, text, and image data represented in tabular format.

\begin{figure*}[t]
\begin{center}
\centerline{\includegraphics[width=1.6\columnwidth]{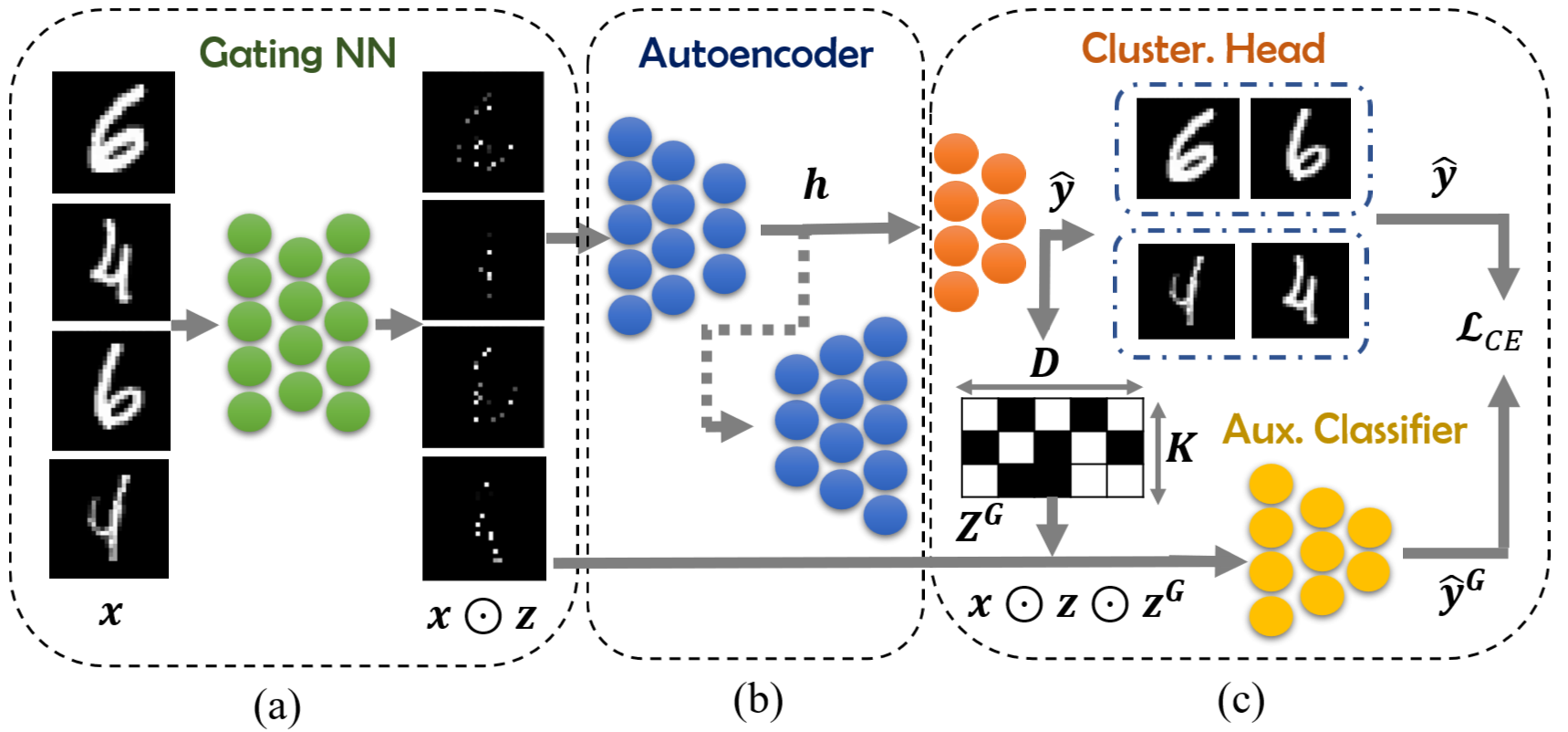}}
\caption{Illustration of the proposed model. The first step involves self-supervision for learning a meaningful latent representation and sample-level informative features.
During this stage, we optimize the parameters of the Gating Network (green) and the autoencoder (blue) that reconstructs $\hat{x}$ from latent embedding $h$. The gating Network learns a sample-specific sparse gate vector $z$ for input sample $x$ such that $x \odot z$ is sufficient for reconstruction via an autoencoder. We train a clustering head (orange) to predict cluster assignments $\hat{y}$ from the latent embedding $h$ by minimizing the mean cluster coding rate loss (see Eq. \ref{eq:clusterhead}). This loss is designed to push clusters apart while making each cluster more compact. The Auxiliary Classifier (yellow) is trained on sparse representations $x \odot z \odot Z^G$ to predict cluster labels and optimizes the cluster level gating matrix $Z^G$. }
\label{fig:explain}
\end{center}
\end{figure*}

This work introduces \textbf{{I}}nterpretable \textbf{{D}}eep \textbf{{C}}lustering ({\textbf{IDC}}), an unsupervised two-stage clustering method for tabular data. The method first selects informative features for reliable data reconstruction, a task that was demonstrated to be correlated with clustering capabilities \citep{song2013auto,han2018autoencoder}. Then, using the sparsified data, we learn the cluster assignments by optimizing neural network parameters subject to a clustering objective function \cite{yu2020learning}.  
In addition, the method provides both instance-level and cluster-level explanations represented by the selected feature set. The model learns instance-level \textit{local gates} that select a subset of features using an autoencoder (AE) trained to reconstruct the original sample. The \textit{global gates} for cluster-level interpretability are derived from the cluster label assignments learned by the model. To ensure sample-level interpretations, the gates are encouraged by the recently proposed discrimination constraint denoted as the total coding rate. Using synthetic data and MNIST, we demonstrate the interpretability quality of our model. Then, we use real-world data from biomed, images, text, and physics to demonstrate that our model can find meaningful clusters while using only a small subset of informative features. In the following sections, we provide a detailed description of our approach.

\section{Related Work}
\paragraph{Unsupervised Feature Selection}
The problem of unsupervised feature selection (UFS) involves identifying variables useful for downstream tasks such as data clustering. Towards this goal, several works have exploited regularized AEs \citep{han2018autoencoder,lee2022self, sokar2022pay, balin2019concrete}, which identify a global subset of features that are sufficient for data reconstruction.
Another line of UFS schemes relies on the graph Laplacian \citep{von2007tutorial} to identify subsets of smooth features with respect to the core structure in the data \citep{he2005laplacian, zhao2012spectral,lindenbaum2021differentiable,shaham2022deep}. Both types of UFS frameworks can help improve downstream clustering capabilities; however, existing global schemes do not provide sample-level or cluster-level interpretability. Although there are recent works on supervised local feature selection \citep{yoon2019invase, yang2022locally} that provide interpretability, we are not aware of any sample-level unsupervised feature selection schemes. Therefore, we present, for the first time, an end-to-end clustering scheme with local feature selection capabilities.

\paragraph{Interpretable Clustering}
\citet{guan2011unified} presented a pioneering work for simultaneous unsupervised feature selection and clustering. The authors proposed a probabilistic model that performs feature selection by using beta-Bernoulli prior in the context of a Dirichlet process mixture for clustering. However, their model can only select cluster-level and dataset-level informative features, whereas our approach offers interpretability with sample-level granularity. In \citep{frost2020exkmc}, the authors proposed tree-based $K$-means clustering as a part of other works in the same direction \cite{lawless2022interpretable, gabidolla2022optimal, cohen2023interpretable, kauffmann2022clustering}. However, the explanations are global only for a given dataset and rely on the whole set of points. In contrast, our approach learns local gates for each sample in the dataset by optimizing a neural network that performs local feature selection, thus producing sample-level interpretations. Since our method is fully parametric, it also offers enhanced generalization capabilities compared to existing schemes. Precisely, our model can predict cluster assignments and informative features for samples not seen during training.

\paragraph{Deep clustering}
Recently, several methods were proposed for NN-based clustering, to name a few: \citep{gao2020deep, niu2021spice, li2022neural, shaham2018spectralnet, cai2022efficient, lv2021pseudo, shen2021you, peng2022xai}. However, these methods primarily focus on vision and rely on domain-specific augmentations and, therefore, can not be applied to tabular datasets. Therefore, we introduce an interpretable NN model for general high-dimensional datasets, such as biomedical tabular data. 

\citet{li2022neural} have generalized the maximum coding rate reduction loss (MCR$^2$) \citep{yu2020learning} for embedding and clustering. The model aims to separate clusters while making them denser. However, this scheme also requires domain knowledge to design semantic preserving augmentations to train the contrastive loss. While such augmentations could be designed effectively for the visual domain, creating such semantic preserving augmentations of tabular data remains an open challenge \citep{qian2023synthcity}. To overcome this limitation, we propose a two-step procedure that involves (i) self-supervision with locally sparse reconstruction that improves interpretability and induces a spectral bias that enhances performance on tabular data, and (ii) an adapted MCR$^2$ \citep{li2022neural} objective for identifying clusters based on diverse features.

\section{Problem Setup}
\label{sec:problem}
We want to cluster a set of data points $\mathbf{X}=\{\myvec{x}_i\}_{i=1}^N$ into $K$ clusters. 
Each data point $\mathbf{x}_i \in R^D$ is a $D$-dimensional vector-valued observation of general type, meaning it does not follow a particular feature structure. Our goal is to learn an 
interpretable clustering model defined by the tuple $\langle f_{\Theta}, \mathcal{S}^{glob} \rangle$ such that $f_{\Theta}(\myvec{x}_i)=\{ \hat{y}_i, \mathcal{S}_i^{loc}\}$ where $ \hat{y}_i \in \{1,2,...,K \}$ is an accurate clustering assignment, and $ \mathcal{S}_i^{loc}$ is a local feature importance set for sample $i$ and defined by $\mathcal{S}_i^{loc}=\{s_i^j \in [0,1]\}_{j=1}^D$. 
$\mathcal{S}^{glob} \in [0,1]^{K \times D}$ is a global feature importance matrix where each gating vector of size $D$ is learned for $K$ clusters. By forcing $|\mathcal{S}_i^{loc}| <<D$, we can attenuate nuisance features and identify (sample-specific) subsets of informative features, thus improving the interpretability of the clustering model. 

Our research is driven by the critical task in biomedicine of cell clustering and identifying marker genes that are associated with each cluster \citep{kiselev2017sc3}. This task involves clustering a high-dimensional dataset while identifying genes with unique patterns in each cluster. Therefore, our goal is to develop an interpretable model that performs well in the clustering task, i.e., can identify groups with semantically related samples while focusing on local subsets of features. We want to highlight that in the supervised setting, this kind of local feature selection has been linked to interpretability by various authors \citep{alvarez2018towards,yoon2019invase,yang2022locally}.

We have extended the concept of supervised interpretability to the unsupervised domain by introducing a new model that combines clustering and local feature selection. This model identifies and eliminates irrelevant input features that do not contribute to the clustering task. Moreover, the model generates unique explanations for each sample and cluster-level interpretations, which can help comprehend the predictions. We have used various metrics to demonstrate that our model enhances interpretability and produces accurate clustering results.

\section{Method}
\label{sec:method}

We suggest a two-step approach involving training self-supervised \textit{local} (sample-specific) gates and latent representations to enhance cluster separation. This approach provides interpretability through local feature selection, which improves performance on tabular data by allowing for the learning of high-frequency prediction functions using Neural Networks (NNs). In contrast to image or audio data, tabular data requires a prediction function with higher frequency components \cite{beyazit2023inductive}. However, the implicit bias of NNs towards learning low-frequency functions, as described in \cite{basri2020frequency}, hinders the performance of vanilla fully connected models on tabular data. We empirically demonstrate that \textit{local} sparsification (gates) can aid NNs in learning high-frequency functions and improving predictive capabilities, as shown in Section \ref{sec:inductive}.

In the second step of our proposed method, we identify the cluster assignments and \textit{global} gates which focus on the driving features at the cluster level. The \textit{global} gates are learned based on the clustering assignments, which are determined by minimizing the \textit{coding rate} (Eq. \ref{eq:clusterhead}) for each cluster. The result of this step is the clustering assignments for each sample, local gates for sample-level interpretation, and \textit{global} gates for cluster-level explanations. The following subsection will present the proposed framework and architecture design in more detail.

\subsection{Local Self Supervised Feature Selection}

Given unlabeled observations $\{\myvec{x}_i\}_{i=1}^N$, where $\myvec{x_i} \in \sR^{D}$, our goal is to learn a prediction function $f_\theta$ (parametrized using a NN) and a feature importance vectors $\myvec{z}_i \in [0, 1]^D$. The feature importance vector will ``highlight'' which subset of variables the model should rely on for clustering each sample $\myvec{x}_i$. This will allow the model to use fewer features for each sample and reduce overfitting when predicting the clusters of unseen samples. 

Towards our goal, we expand upon the recently proposed stochastic gates \citep{lindenbaum2021differentiable, yamada2020feature,jana2023support} by allowing them to learn the importance vector locally (for each sample). These stochastic gates are continuously relaxed Bernoulli variables 
defined based on the hard thresholding function for feature $d$ and sample $i$: 
\begin{equation}\label{eq:stg}
{z}^d_i = \max(0,
\min(1,0.5+ {\mu}^d_i + {\epsilon}^d_i)),
\end{equation}
where ${\epsilon}^d_i$ is drawn from $\mathcal{N}(0 ,\sigma^2)$. The value ${\mu}^d_i$ is the logit output of the network before being passed through the hard thresholding function (in Eq. \ref{eq:stg}). In our model, $\sigma$ is fixed to $0.5$ throughout training, as suggested in \citep{yamada2020feature}. The injected noise is controlled by $\sigma$ and helps push the converged values of ${z}^d_i$ towards 0 or 1. For more information, please see \citep{yamada2020feature} and \citep{yang2022locally}.
The parameters ${\myvec{\mu}_i}\in \mathbb{R}^D,i=1,...,N$ are specific to each sample and are predicted using a gating network $f_{\theta_1}$. In other words, ${\myvec{\mu}_i}=f(\myvec{x}_i|\theta_1)$, where $\theta_1$ are the weights of the gating network. These weights are learned simultaneously with the weights of the prediction network $\theta_2$ by minimizing the following loss:
\begin{equation*}
\label{eq:sparse_loss}
\mathcal{L}_{\text{sparse}}=\mathbb{E}_{\myvec{z}_i, \myvec{x}_i}\big[ {\cal{L}} (f_{\theta_2} (\myvec{x}_i\odot\myvec{z}_i)) + \lambda \cdot \mathcal{L}_{\text{reg}}(\myvec{z}_i) \big],
\end{equation*}
where ${\cal{L}}$ is a desired prediction loss, for example, a clustering objective function or reconstruction error. The Hadamard product is simply an element-wise multiplication and is denoted by $\odot$. We compute the empirical expectation over $\myvec{x}_i$ and $\myvec{z}_i$, where $i$ is an index that ranges over a dataset of size $N$. The term ${\cal{L}}_{\text{reg}}(\myvec{z}_i)$ is a regularizer designed to sparsify the gates. It is defined as follows: ${\cal{L}}_{\text{reg}} = \|\myvec{z}_i\|_0$. After taking the expectation over $\myvec{z}_i$ and the samples $\myvec{x}_i$, $\mathbb{E}[{\cal{L}}_{\text{reg}}]$ can be rewritten using a double sum in terms of the Gaussian error function ($\erf$):
\begin{equation}
    {\cal{L}}_{\text{reg}} = \frac{1}{N}\sum^N_{i=1} \sum^D_{d=1}\left(\frac{1}{2} - \frac{1}{2} \erf\left(-\frac{\mu^{d}_i + 0.5}{\sqrt{2}\sigma}\right) \right),
\label{eq:reg}
\end{equation}
here, we calculate the expectation using the parametric definition of $\myvec{z}_i$.

We have chosen a denoising autoencoder \citep{vincent2008extracting} for our prediction network. It aims to identify and select only the informative features required for reconstruction by disregarding nuisance features. The network is trained through self-supervision using a reconstruction loss with domain-agnostic augmentations. This helps the network learn a latent embedding of the input sample and prompts the gating network to open only the gates needed to reconstruct data. The model consists of the following components:
\begin{itemize}
\item \textit{Gating Network}: \quad 
$f_{\theta_G}(\myvec{x}_i)=\myvec{z}_i$, is a hypernetwork that predicts the gates $\myvec{z}_i$ vector for sample $\myvec{x}_i$, where $\myvec{z}_i \in[0,1]^D$ (each element is defined based on Eq.\ref{eq:stg}).

\item \textit{Encoder}: \quad  
$f_{\theta_E}(\myvec{x}'_i)=\myvec{h}_i$, is a mapping function that learns an embedding $\myvec{h}_i$ based on the element-wise gated sample $\myvec{x}'_i =\myvec{x}_i \odot \myvec{z}_i $.

\item \textit{Decoder}: \quad  $f_{\theta_D}(\myvec{h}_i)=\hat{\myvec{x}}_i$, that reconstructs $\myvec{x}_i$ based on the embedding $\myvec{h}_i$.
\end{itemize}
We utilize an autoencoder with parameters $\theta_E \cup \theta_D $ and employ gated input reconstruction loss $\mathcal{L}_{\text{recon}}( f_{\theta_D}(f_{\theta_E} (f_{\theta_G} (\myvec{x}_i) \odot \myvec{x}_i)), \myvec{x}_i)$ to measure the deviation between estimated $\hat{\myvec{x}_i} = f_{\theta_D}(f_{\theta_E} (f_{\theta_G} (\myvec{x}_i) \odot \myvec{x}_i))$ and input sample $\myvec{x}_i$. We introduce input \citep{vincent2008extracting} and latent data augmentations \citep{doi2007robust} to learn semantically informative features. Additional details about these augmentations are in Appendix \ref{app:augmentations}. Additionally, we introduce an extra \textit{gates total coding rate loss}, $\mathcal{L}_{\text{gtcr}}(\myvec{Z})$ that encourages the selection of unique gates for each sample. The following equation defines this loss: 
\begin{equation*}
\label{eq:tcr}
 \mathcal{L}_{\text{gtcr}} = - 
 \frac{1}{2} \cdot \text{logdet} (\mathbf{I} + \frac{D}{N_B \cdot \epsilon} \cdot (\myvec{Z}^T\myvec{Z})),
\end{equation*} 
which is approximately the negative Shannon coding rate of a multivariate Gaussian distribution \citep{yu2020learning} up to precision $\epsilon$, and is defined for a mini-batch of size $N_B$ of normalized local gates $\myvec{Z} \subseteq \{\myvec{z}_i \}_1^N$. 
This component, which is inspired by \citep{li2022neural}, aims to decrease the correlation between gate vectors $\myvec{z}_i$.

The final loss for predicting sparse data is calculated as follows: 
\begin{equation}
    \mathcal{L}_{\text{sparse}}=\mathcal{L}_{\text{recon}} +\mathcal{L}_{\text{gtcr}} + \lambda \cdot \mathcal{L}_{\text{reg}}.
    \label{eq:loss}
\end{equation}
The weight term for regularization, represented by $\lambda$, is gradually increased in training using a cosine function scheduler. The autoencoder's weights are initially set through standard reconstruction objective training. In Section \ref{sec:ablation}, we conduct an ablation study to validate the importance of each component of our loss.

Our model is designed to minimize the number of features required for data reconstruction. This is achieved by sparsifying the input samples. Additionally, the gated input reconstruction loss helps the model learn local masks, which are used to attenuate noisy features and improve interpretability. The gates' total coding rate loss encourages the selection of diverse gates across samples, ensuring that informative features are not constant across the data.

\subsection{Cluster assignments with Global 
Interpretations}

\paragraph{Clustering Head}

During the clustering phase, our goal is to compress the learned representations of gated samples of $\mathbf{X}$ into $K$ clusters and predict cluster-level gates. To accomplish this, we train a clustering head $f_{\theta_C}(\myvec{h}_i)=\hat{y}_i$, which learns cluster one-hot assignments $\hat{y}_i\subset \{1,...,K\}$. The model outputs logits values $\myvec{\pi}_i^{1 \times K}$ for each sample. These logits values are then converted to cluster assignment probabilities $\myvec{\hat{y}_i}$ by using a Gumbel-Softmax \citep{jang2016categorical} reparameterization:
    \begin{equation*}
    \hat{y}_i^k = \frac{\text{exp}((\log({\pi}_i^k)+g_i^k)/\tau)}{\sum_{j=1}^K \text{exp}((\log({\pi}_i^k)+g_i^k)/\tau)}, 
   \qquad \text{for } k=1, ..., K.
    \end{equation*}
where $\myvec{g}_i^{1 \times K}$ are i.i.d. samples drawn from a Gumbel(0, 1) distribution. The variable $\tau$ represents a temperature hyperparameter, where larger values of $\tau$ produce a uniform distribution of $\hat{\myvec{y}}$. On the other hand, decreasing the temperature leads to one-hot vectors. This neural network is optimized with the following loss \begin{equation}\label{eq:clusterhead}
     \mathcal{L}_{\text{head}} =  \sum_{k=1}^K \frac{1}{2} \cdot \text{logdet} \big[\mathbf{I} + \frac{d_h}{N_B \cdot \epsilon} \cdot (\myvec{H}_k^{T} \myvec{H}_k)\big],
\end{equation}
where $\mathbf{H}_k = \{ \myvec{h}_{i} : f_{\theta_C}(\myvec{h}_i)=k\} \subseteq \{ \myvec{h}_i = f_{\theta_E}(\myvec{x}_i \odot \myvec{z}_i)\}_{i=1}^N$ is a mini-bach of size $N_B$ embedding vectors for all samples $\myvec{x}_i$'s which were assigned to cluster $k$. $d_h$ is the embedding dimension and $\epsilon$ is the hyperparameter that denotes the desired precision of coding rate \cite{yu2020learning}. The aim is to decrease the average coding rate for each cluster of embeddings $\mathbf{H}_k$ to make them more compact.

\paragraph{Global Interpretations}
\vskip -0.1in
To provide global cluster-level interpretations, we develop a gate matrix called $\mathbf{Z}_{G}\in\{0,1\}^{K \times D}$. Each row in the matrix corresponds to a cluster and each column to an input variable. We use an \textit{auxiliary classifier}, $f_{\theta_{AC}}$, to train this gate matrix. The classifier accepts a gated representation of $\myvec{x}_i$, defined by $\myvec{x}_i \odot \myvec{z}_i \odot \myvec{z}_{G}^k$, where $\myvec{z}_{G}^k$ is a global gates vector learned for cluster $k=\text{argmax}_{1,...,K} \myvec{\hat{y}}_i$. The locally sparse samples $\myvec{x}_i \odot \myvec{z}_i$ learned with the autoencoder are multiplied by global gates $\textbf{Z}_G$ and passed through a single-hidden-layer classifier. This classifier is trained to output cluster assignments $\hat{\myvec{y}}_i$ identical to those predicted with clustering head $f_{\theta_C}$ using a cross-entropy loss. During inference, only $\mathbf{Z}_{G}$ is used, and $f_{\theta_{AC}}$ is not needed.

In our method, we use a regularization loss term called $\mathcal{L}_{\text{reg}}$ with an increasing weight $\lambda_g$ to sparsify the gates in the global gates matrix. This is added to the clustering loss term $\mathcal{L}_{clust}$, which is the sum of the clustering-head term $\mathcal{L}_{\text{head}}$ and the cross-entropy loss term $\mathcal{L}_{CE}$. To summarize, we train the clustering head to predict assignments along with global gates. We use an auxiliary classifier to optimize the global gates while being trained in a self-supervised manner on the pseudo labels predicted by the clustering head.

\section{Interpretability}
\label{sec:interpret}
Practitioners may need interpretability at different levels of granularity. At a coarser level, it is useful to identify which features are common to a group of semantically related samples (or clusters) \citep{guan2011unified}. At a finer level of detail, we aim to find unique explanations for each data point. Specifically, we want to know which features drive the model to make specific predictions \citep{alvarez2018towards}.

Although the internal workings of a model may remain a black box to the user, the relationship between the input features and the model's predictions can provide some insight into its interpretability. Recently, several criteria have been proposed to assess the interpretability of supervised models \citep{alvarez2018towards,yang2022locally}, as outlined in the following paragraphs. We will also discuss how these criteria can be adapted for an unsupervised setting.

 \paragraph{Diversity}
 \vskip -.05in

We expect a good interpretability model to identify different sets of variables as driving factors for explaining distinct
clusters. This is measured by the \textit{diversity} metric, which is calculated by finding the negative mean Jaccard similarity between cluster-level informative features across all pairs of clusters. In simple terms, given a set of indices $S_{c_{i}}\subset 1,...,D$ that indicate the selected informative variables of cluster $c_i$, where $i=1,...,K$, the \textit{diversity} is defined as $1-\sum_{i \neq j} \frac{J(S_{c_i}, S_{c_j})}{K \cdot (K-1) / 2} $. Here, $J$ is the Jaccard similarity between two sets. A score of 1 is obtained when there is no overlap between cluster-level features, indicating perfect diversity.

\paragraph{Faithfulness}
\vskip -.05in
An interpretation is faithful if it accurately represents the reasoning behind the model's prediction function. To evaluate this quantity, \textit{faithfulness} measures the correlation between the predictivity of the model and the feature importance. To calculate this, we first compute a feature's importance value, for instance, the value of our corresponding predicted gate averaged over all samples. We then sort the vector of importance values, remove features individually starting from the most important one, and measure the clustering model's performance. If the model's performance decreases in a monotonic manner with the importance of the removed features, we will get a high correlation, indicating that the model's prediction is faithful to the learned feature importance values. An example of this metric is presented in Fig. \ref{fig:overlay}.

 \paragraph{Uniqueness}
\vskip -.05in

Since we are interested in sample-level interpretation, we extend the \textit{diversity} to a metric that compares samples instead of clusters. Specifically, we propose to measure the \textit{uniqueness} of the selected features for similar samples, or in other words, how granular our explanations are. 
We define \textit{uniqueness} of an explanation $\myvec{z}_i$ by:
 $\frac{1}{|I_i|} \sum_{\ell \in I_i} {\min} \frac{\|\myvec{z}_i - \myvec{z}_{\ell} \|_2}{ \|\myvec{x}_i-\myvec{x}_{\ell} \|_2}$, where $I_i$ is the set of indices of the $r$ nearest neighbors of $\myvec{x}_i$, and $\myvec{z}_i,\myvec{z}_{\ell}$ are feature weights for samples $\myvec{x}_i,\myvec{x}_{\ell}$. The smaller this value is, the less sample-specific the interpretation of the model is. Therefore, we want our model to obtain high \textit{uniqueness} values. In case the uniqueness is not a desired property and could raise confusion in the produces interpretations, the loss term $\mathcal{L}_{\text{gtcr}}$ in Eq. \ref{eq:loss} should be removed.

\paragraph{Generalizability}
\vskip -.05in
We want the interpretation of the prediction to generalize to other simple prediction models. Specifically, if the selected features also lead to high predictive capabilities across different models, this may indicate that the interpretability is not an artifact of a specific model architecture or training instance. To evaluate this quantity, termed \textit{generalizability}, we follow \citep{yang2022locally} and measure the accuracy of SVM when applied to the data, masked by the most informative features as identified by each method.

\section{Experiments}

We performed six different types of experiments in our study. Firstly, we tested our method's clustering and local feature selection capabilities on a synthetic dataset. Then, we evaluated the interpretability of our model using MNIST. The main experiment focused on assessing our method's ability to cluster real-world tabular datasets, including those with small sample sizes, high-dimensional biomed datasets, and physics and text datasets. It's worth noting that the real-world datasets we used are still considered challenging by several clustering studies that have used tabular data \citep{shaham2022deep, xu2023efficient}.

In the fourth experiment, we demonstrated that our model can also perform well on image data without needing domain-specific augmentations. The datasets we used for our experiments have been summarized in Table \ref{app:datasets} in the Appendix. In the fifth experiment, we showed that our local gates help the network learn a prediction function that includes high-frequency components. Finally, we conducted an ablation study to demonstrate the importance of each component in our model.

\begin{table*}[t!]
 \vspace{-0.15 in}
    \centering
    \caption{Evaluating the interpretability quality of our model on the $\text{MNIST}_{10K}$ data. Our IDC model improved clustering accuracy compared to baselines that do not use semantic-preserving image augmentations. We focus on the $|\mathcal{S}|=15$ most informative features as provided by each interpretation model. 
    We compare the interpretability of our model, as predicted by the gates (IDC), to (i) the top features explained by SHAP trained based on a $K$-means model, (ii) Integrated Gradients, and (iii) Gradient SHAP applied as explainers to our model and TELL. Our local gating network is generalizable (88.5) by selecting faithful (0.96) and unique (0.69) features while providing comparable diversity values (94.8).}
    \begin{adjustbox}{width=1.7\columnwidth,center}

                 \begin{tabular}{|l|l|c|c|c|c|c|c|}
                \hline 
              Method & ACC $\uparrow$ & $|\mathcal{S}| \downarrow$ & Uniqueness $\uparrow$ 
              & Diversity $\uparrow$ &  Faithfulness $\uparrow$ & Generalizability $\uparrow$ \\ 
                 \hline 
                $K$-means + SHAP & 53.34 & 15 & 0.12 
                & \textbf{100.0} & 0.79 & 29.1 \\ \hline

                TELL + IntegGrads & 74.79 & 15 & 0.03 
                & 89.1 & 0.67 & 75.0\\

                TELL + GradSHAP & 74.79 & 15 & 0.15 
                & 92.5 & 0.63 & 78.9 \\ \hline
                 
                IDC w/o gates + IntegGrads & 82.32 & 15 
                & 0.02 
                & 95.8 & 0.78 & 80.3\\
                
                IDC w/o gates + GradSHAP & 82.32 & 15 & 0.08 
                & \textbf{100.0} & 0.86 & 59.9\\
                \hline
                IDC + IntegGrads & \textbf{83.45} & 15 
                & 0.01 
                & 95.3 & 0.94 & 63.7\\

                 IDC + GradSHAP & \textbf{83.45} & 15 & 0.02 
                 & 97.0  &  0.93 & 66.0\\ \hline  
                 \rowcolor[HTML]{E8D9FB} IDC & \textbf{83.45} & 15 & \textbf{0.69} 
                 & 94.8 & \textbf{0.96} & \textbf{88.5}\\
                
                 \hline
         \end{tabular}
    \end{adjustbox}
    \label{tab:explain}
\end{table*}

\subsection{Evaluation}

\paragraph{Interpretability Quality}

We use the metrics detailed in Section \ref{sec:interpret} to compare the interpretability of our method to the popular SHAP feature importance detection method \citep{lundberg2017unified} implemented here \footnote{https://github.com/slundberg/shap}. In addition, we compare Gradient SHAP \footnote{https://captum.ai/api/gradient\_shap.html} and Integrated Gradients \citep{sundararajan2017axiomatic}, both trained on IDC clustering model predictions. Additionally,  both Gradient SHAP and Integrated Gradients explainers are evaluated while being applied on the TELL \cite{peng2022xai} model predictions. For uniqueness, we set the number of neighbors to 2 ($r=2$).
\paragraph{Clustering Accuracy}
We use three popular clustering evaluation metrics: Clustering Accuracy (ACC), The adjusted Rand index score (ARI), and Normalized Mutual Information (NMI).
\subsection{Results}
\paragraph{Synthetic Dataset}

We started by testing our model's performance with a synthetic dataset inspired by \citep{armanfard2015local}. The dataset includes three informative features, denoted as $X[j] \in [-1,1],j=1,..,3$, in which the samples are generated using Gaussian blobs \footnote{https://scikit-learn.org/stable/modules/generated/
sklearn.datasets.make\_blobs.html}. We added ten background features, resulting in a total of 13 features. 

\begin{figure}[!t]
\begin{center}
\includegraphics[width=.49\columnwidth]{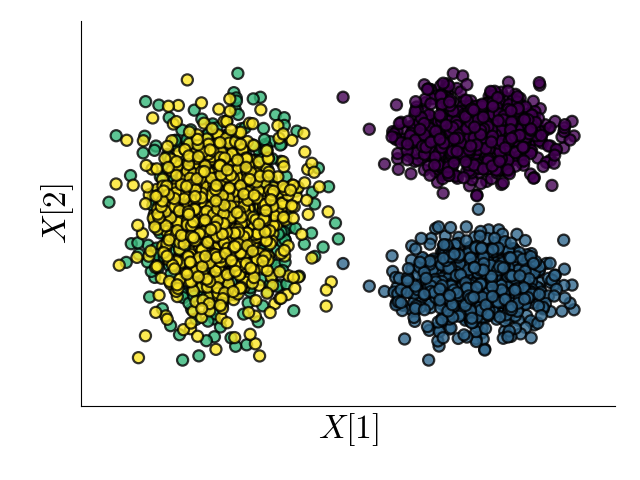}
\includegraphics[width=.49\columnwidth]{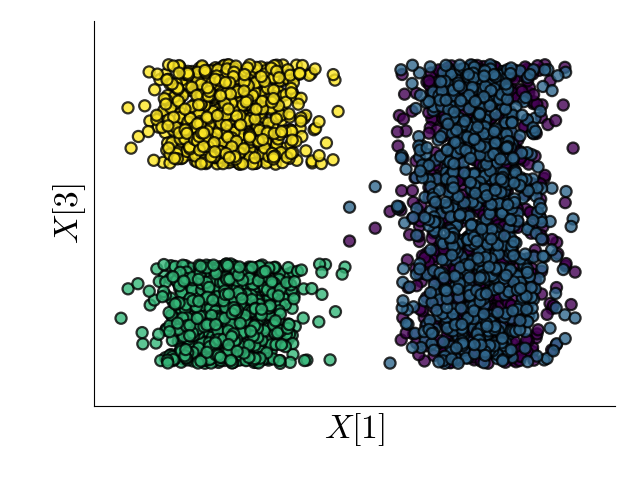}
\caption{Visualization of the synthetic dataset. To separate between clusters, the model should select one of the pairs $\{X[1]
, X[2]\}$ or $\{X[1], X[3]\}$ of non-background features.}
\label{fig:synth}
\end{center}
\end{figure}

The samples are equally distributed between 4 clusters, with $\sim800$ samples in each cluster. Given the first two dimensions $\{X[1], X[2]\}$, only 3 clusters are separable, and the same property holds for dimensions pair $\{X[1], X[3]\}$. A visualization of these leading features appears in Figure \ref{fig:synth}. A good interpretable clustering model should correctly identify the four clusters while selecting the informative features for each cluster. For a more detailed explanation of how the dataset was generated, please refer to Appendix \ref{app:synthgen}.

We compute clustering accuracy (ACC) and F1-score to assess the model's effectiveness, which measures the feature selection quality. Since we know which are the informative features for each sample, we can determine the precision and recall for gate-level feature selection. We compare our results with those of $K$-means with informative features identified by SHAP. Our method leads to 99.91\% clustering accuracy, thus significantly outperforming $K$-means with 25.72\%. Moreover, IDC yields an average F1-score of 88.95 (compared to 49.65 by $K$-means), demonstrating its ability to identify the correct informative features. 

\paragraph{Interpretability Results on MNIST$_{10K}$}
\vskip -0.1in
Our next interpretability evaluation involves $\text{MNIST}_{10K}$ images, with $1K$ images for each category. We train our model with and without (w/o) gates and evaluate its clustering accuracy. These are compared to  TELL \cite{peng2022xai}, an interpretable deep clustering model, and to $K$-means. As demonstrated in Table \ref{tab:explain}, our model with the gates leads to improved clustering capabilities. It is important to note that we do not use any domain-specific image augmentations and that existing schemes that do use will typically lead to higher clustering accuracies on this data.

\begin{figure*}[ht]
\begin{center}
\includegraphics[width=0.9\columnwidth]{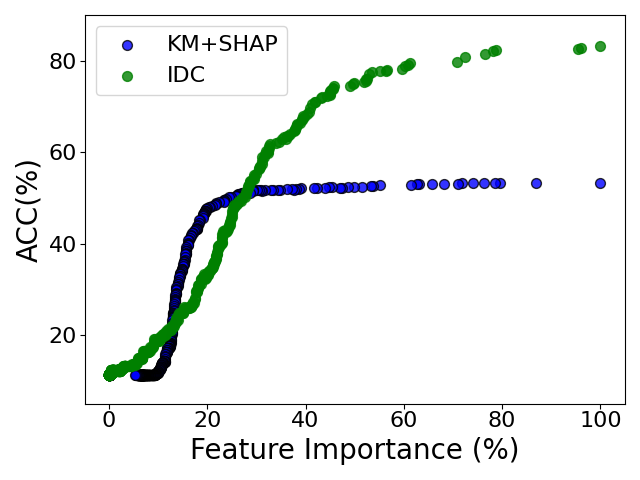}
\includegraphics[width=1.1\columnwidth]{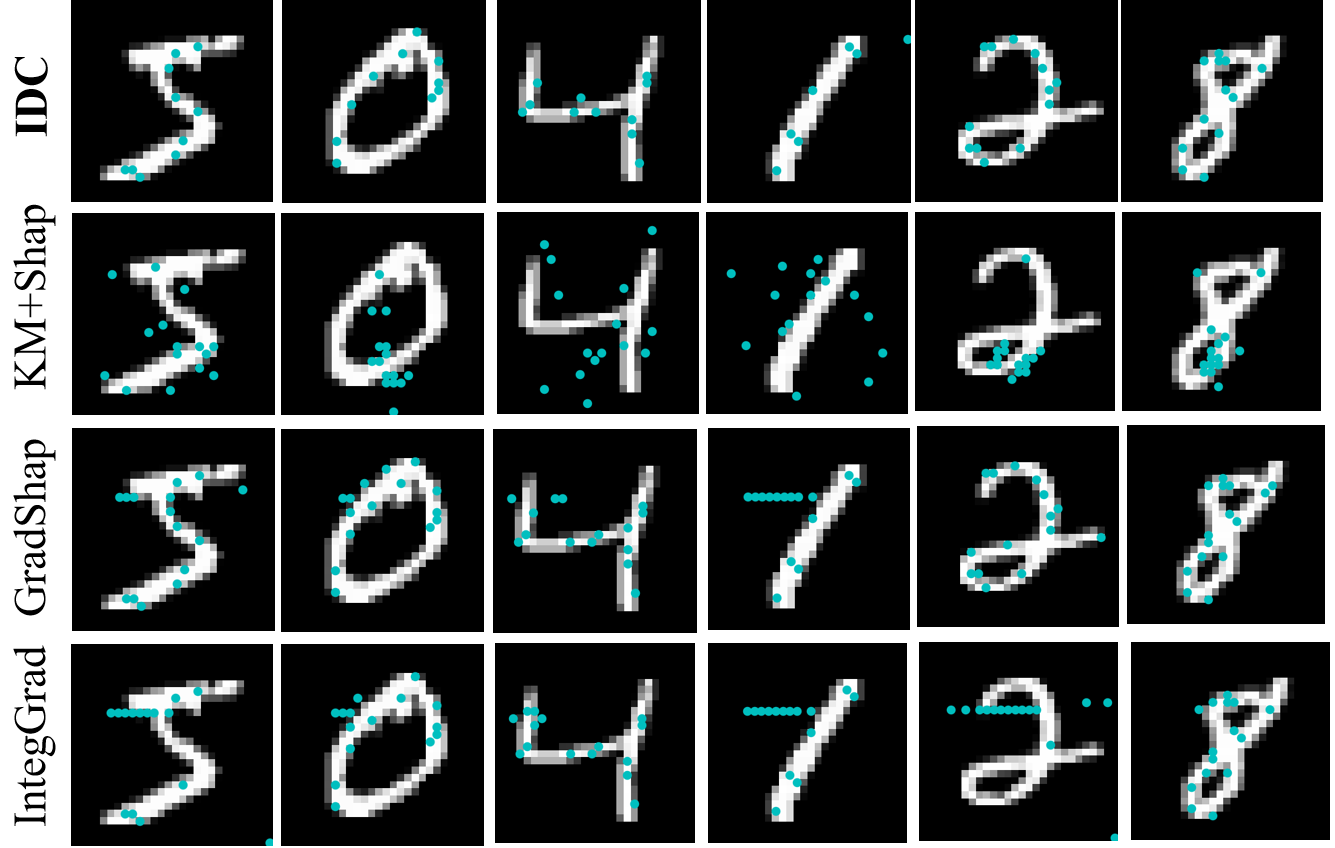}
\end{center}
\caption{ \textbf{Left}: Faithfulness plot of the proposed method (green) and $K$-Means+SHAP (blue) on MNIST$_{10K}$ subset. Accuracy drop and feature importance are well correlated for our approach $0.96$ (see green dots) while less correlated for SHAP features with $K$-means clustering $0.79$ (see blue dots). Furthermore, notice that $K$-means accuracy only reaches $53\%$ while our method $83\%$. 
\textbf{Right}: Features selected by different interpretability model using $\text{MNIST}_{10K}$. The features are learned during clustering training as proposed by our approach (top), features selected by SHAP with $K$-means predictor (KM-SHAP), features obtained from Gradient-SHAP (GradSHAP), and features selected by Integrated Gradients explainer (bottom).
}
\label{fig:overlay}
\end{figure*}

In terms of interpretability, we compare the most informative features identified by each method based on the metrics described in Section \ref{sec:interpret}. As indicated in Table \ref{tab:explain},
IDC with our gates (bottom row) improves \textit{faithfulness}, \textit{uniqueness}, and \textit{generalizability} while maintaining competitive \textit{diversity}. In the left panel of Fig. \ref{fig:overlay}, we plot the model's accuracy against the feature importance. Providing a more detailed view on the \textit{faithfulness} of the interpretation. Finally, in the right panel of this figure, the top 15 features selected by each model. It can be seen that \textbf{IDC} selects more informative features that are local for each sample.

\begin{table*}[ht]
\caption{Clustering Accuracy on Real Datasets. We compare the proposed method against leading unsupervised feature selection methods followed by $K$-means clustering, the interpretable deep clustering model TELL \cite{peng2022xai} and recently published k-DVAE \cite{caciularu2022entangledmixture}.  Our model produces superior results on 9 of 11 datasets and second-rank accuracy on two datasets.
}
\label{tab:real1}
\begin{adjustbox}{width=2.05\columnwidth,center}
\begin{tabular}{|r|lllllllllll|}
\toprule
\textbf{Method}$/$\textbf{Data} & TOX-171 & ALLAML & PROSTATE & SRBCT & BIASE & INTESTINE & PBMC-2 & CNAE-9 & MFEATZ &  MiniBooNE & ALBERT \\ 
\midrule
KM  & 41.5 $\pm$ 2 & 67.3 $\pm$ 3 & 58.1 $\pm$ 0 & 39.6 $\pm$ 3 & 41.8 $\pm$ 8 & 54.8 $\pm$ 3 & 52.4 $\pm$ 0 & 44.9 $\pm$ 5 & 56.2 $\pm$ 0 & 71.6 $\pm$ 0 &  55.7 $\pm$ 0 \\
LS & 47.5 $\pm$ 1 & 73.2  $\pm$ 0 & 58.6 $\pm$ 0 & 41.1 $\pm$ 3 & 83.8 $\pm$ 0 & 43.2 $\pm$ 3 & 57.2 $\pm$ 0 & 35.9 $\pm$ 3 & 56.6 $\pm$ 3 &  71.9 $\pm$ 0 & 64.2 $\pm$ 0\\
MCFS & 42.5 $\pm$ 3 & 72.9 $\pm$ 2 & 57.3 $\pm$ 0 & 43.7 $\pm$ 3 & 95.5 $\pm$ 3 & 48.2 $\pm$ 4 & 60.6 $\pm$ 12 & 43.4 $\pm$ 9 & 39.4 $\pm$ 9 &  71.6 $\pm$ 0 & 60.9 $\pm$ 0\\
SRCFS & 45.8 $\pm$ 6 & 67.7 $\pm$ 6 & 60.6 $\pm$ 2 & 33.5 $\pm$ 5 & 50.8 $\pm$ 5 & 58.1 $\pm$ 10 & 58.5 $\pm$ 0 & 34.3 $\pm$ 3 & 58.5 $\pm$ 1 &  71.6 $\pm $ 0 & 56.8 $\pm$ 0\\
CAE & 47.7 $\pm$ 1 & 73.5 $\pm$ 0 & 56.9 $\pm$ 0 & \textbf{62.6} $\pm$ 7 & 85.1 $\pm$ 2 & 51.9 $\pm$ 3 & 59.1 $\pm$ 6 & 45.3 $\pm$ 2 & 70.0 $\pm$ 0 & 71.6 $\pm$ 0 & 64.1 $\pm$ 0 \\ 
DUFS  & 49.1 $\pm$ 3 & 74.5 $\pm$ 1 & 64.7 $\pm$ 0 & 51.7 $\pm$ 1  & \textbf{100} $\pm$ 0 & 71.9 $\pm$ 7 & 57.6 $\pm$ 9 & 46.3 $\pm$ 0 & 57.3 $\pm$ 9 & 71.6 $\pm$ 0 & 62.3 $\pm$ 0 \\ 
TELL & 28.7 $\pm$ 3 & 66.7 $\pm$ 14 & 63.6 $\pm$ 0 & 38.9 $\pm$ 8 &84.6 $\pm$ 2  & 52.1 $\pm$ 28 &  52.1 $\pm$ 1 & 11.1 $\pm$ 0 & 63.8 $\pm$ 36 &  75.6 $\pm$ 5 & 60.6 $\pm$ 3 \\

k-DVAE & 49.1 $\pm$ 5 & 72.6 $\pm$ 5 & 59.5 $\pm$ 2 & 49.4 $\pm$ 12  & 39.3 $\pm$ 6 & 55.7 $\pm$ 7 & 50.1 $\pm$ 0 & 62.8 $\pm$ 7 & 66.5 $\pm$ 5 & 71.7 $\pm$ 0 & 51.5 $\pm$ 1 \\

\rowcolor[HTML]{E8D9FB} \textbf{IDC} & \textbf{50.6} $\pm$ 3 & \textbf{77.5} $\pm$ 6 & \textbf{65.3} $\pm$ 3 & 55.4 $\pm$ 5 & 95.7 $\pm$ 1 & \textbf{74.2} $\pm$ 2 &\textbf{65.1} $\pm$ 5 & \textbf{66.0} $\pm$ 7 & \textbf{86.81} $\pm$ 4 & \textbf{77.0} $\pm$ 4 & \textbf{64.2} $\pm$ 4\\
\midrule

\end{tabular}
\end{adjustbox}
\end{table*}

\paragraph{Real Tabular Data}

We conducted a benchmark of our method using 11 real tabular datasets. Most of these datasets are from different biological domains, such as Tox-171, ALLAML, PROSTATE, SRBCT, BIASE, INTESTINE, and PBMC-2. These datasets usually have more features than samples, making it challenging for clustering models to predict cluster assignments accurately. Additionally, we used image (MFEATZ), text (CNAE-9, ALBERT) and physical measurements (MiniMBooNE) datasets, all of which were treated as tables.

Since most datasets are tabular and high dimensional, we compare our clustering capabilities to several unsupervised feature selection (UFS) models, which have been demonstrated to lead to state-of-the-art clustering results on these datasets \cite{lindenbaum2021differentiable}. Here, we follow the evaluation protocol of \cite{lindenbaum2021differentiable} when using the UFS methods.
Table \ref{tab:real1} presents clustering accuracy of different methods: $K$-means on the full set of features (KM), TELL an interpretable deep clustering model \cite{peng2022xai}, variational autoencoder based k-DVAE \cite{caciularu2022entangledmixture} and different leading UFS models followed by $K$-means. These include SRCFS \citep{huang2019unsupervised}, Concrete Autoencoders (CAE) \citep{abid2019concrete}, DUFS \citep{lindenbaum2021differentiable}, and our model IDC. This table shows that our model leads to the best or second-best clustering accuracy across all datasets. We run every model 10 times for each dataset and report the mean accuracy with standard deviation.

\paragraph{Extension to Image Domain}
We conducted usability tests on our model using popular image datasets. Initially, we assessed the performance of IDC both with and without the gates on MNIST and FashionMNIST. The results, as presented in Table \ref{tab:images}, indicate that the gating network improves the model's ability to cluster data accurately and select only a few relevant features, thereby enhancing interpretability. 

In our second experiment, we utilized CIFAR10 and compared our model to other deep clustering models that do not utilize image augmentations. These models include TELL \citep{peng2022xai}, VaDE \citep{jiang2016variational}, and DEC \cite{xie2016unsupervised}. The results of the experiment are presented in Table \ref{tab:exp:cifar}. Our model produced competitive results while only selecting approximately 20\% of the input features. It is worth noting that while these results are not state-of-the-art in deep image clustering, they were achieved without using strong image augmentations like resizing, translation, or rotations. Such augmentations are incompatible with our gating procedure, as the informative features may vary from one augmentation to the next. Additionally, these augmentations do not apply to tabular data..

\begin{table}[t]
\caption{Clustering evaluation and the number of features selected (last column). IDC with the proposed gates improves clustering accuracy by using only $\sim 16$ from $\text{MNIST}_{60K}$ and  $\sim 69$ features from FashionMNIST datasets.}
 \label{tab:images}
\begin{adjustbox}{width=0.99\columnwidth,center}
\begin{tabular}{|l|c|cccc|}
\hline
Dataset & Method & ACC $\uparrow$ & ARI $\uparrow$ & NMI $\uparrow$ & $|\mathcal{S}|$ $\downarrow$\\ 

\hline

$\text{MNIST}_{60K}$  & IDC (w/o gates) & 81.1 & 75.9 & 80.3 & 784\\
&  \cellcolor[HTML]{E8D9FB} IDC  &  \cellcolor[HTML]{E8D9FB}\textbf{87.9} &  \cellcolor[HTML]{E8D9FB}\textbf{82.8} &  \cellcolor[HTML]{E8D9FB}\textbf{85.1} &  \cellcolor[HTML]{E8D9FB} \textbf{15.81} \\
\hline
FashionMNIST  & IDC (w/o gates) & 61.0 & \textbf{49.3} & 62.7 & 784\\
&   \cellcolor[HTML]{E8D9FB} IDC &  \cellcolor[HTML]{E8D9FB} \textbf{61.9}&  \cellcolor[HTML]{E8D9FB} 49.1 &  \cellcolor[HTML]{E8D9FB} \textbf{63.3}&  \cellcolor[HTML]{E8D9FB} \textbf{68.6}\\
\hline
\end{tabular}
\end{adjustbox}
 \caption{Clustering performance on CIFAR10 dataset. IDC model selects 586 features (on average) out of 3,072.}
\label{tab:exp:cifar}
 \begin{adjustbox}{width=0.6\columnwidth,center}
\begin{tabular}{|l |c |c| c |}
 \hline
Model  & ACC $\uparrow$ & ARI $\uparrow$ & NMI $\uparrow$ \\
\hline
TELL   & \textbf{25.65} & 5.96	 & 10.41 \\ 
VaDE 	 & 20.87 & 3.95 & 7.20  \\ 
DEC & 18.09 &  2.47 & 4.56 \\
\rowcolor[HTML]{E8D9FB} IDC      & 25.01 & \textbf{6.16} & \textbf{11.96}  \\
\hline 
\end{tabular}
\end{adjustbox}

\label{tab:ablation}
\caption{Ablation study on $\text{MNIST}_{60K}$ dataset}
\begin{adjustbox}{width=0.99\columnwidth,center}
\begin{tabular}{|l|c|c|c|}
\hline
Model & ACC $\uparrow$ & ARI $\uparrow$ & NMI $\uparrow$ \\
\hline
\rowcolor[HTML]{E8D9FB} IDC & $\mathbf{87.9} $ & $\mathbf{82.8}$& $\mathbf{85.1}  $ \\

IDC w/o $\mathcal{L}_{\text{reg}}$ & $85.9$ \color{red}{(-2.0)} & $81.2 $\color{red}{(-1.6)} & $84.7$ \color{red}{(-0.4)} \\
IDC w/o latent denoising & $86.5$ \color{red}{(-1.4)} & $80.9$ \color{red}{(-1.9)} & $83.2 $ \color{red}{(-1.9)} \\
IDC w/o input denoising & $84.3$ \color{red}{(-3.6)} & $80.0$  \color{red}{(-2.8)} & $83.9$ \color{red}{(-1.2)}  \\
IDC features + $K$-Means & $65.5$ \color{red}{(-22.4)} & $49.3$ \color{red}{(-33.5)}& $57.6$ \color{red}{(-27.5)}\\
IDC w/o $\mathcal{L}_{\text{recon}}$ & $18.0$ \color{red}{(-69.9)}& $2.6$ \color{red}{(-80.2)} & $4.3$ \color{red}{(-80.8)}\\
\hline
\end{tabular}
\end{adjustbox}
\end{table}

\subsection{Sparsity and Inductive Frequency Bias}
\label{sec:inductive}

According to a study by \citep{beyazit2023inductive}, tabular datasets generally require higher frequency target functions than images. However, several authors have shown that neural networks tend to learn low-frequency functions faster than higher ones. This might explain why tree-based models often outperform neural networks. For more information, please refer to the appendix section on inductive learning (Section \ref{app:inductive}).

\begin{figure}[!t]
\begin{center}
\includegraphics[width=.89\columnwidth]{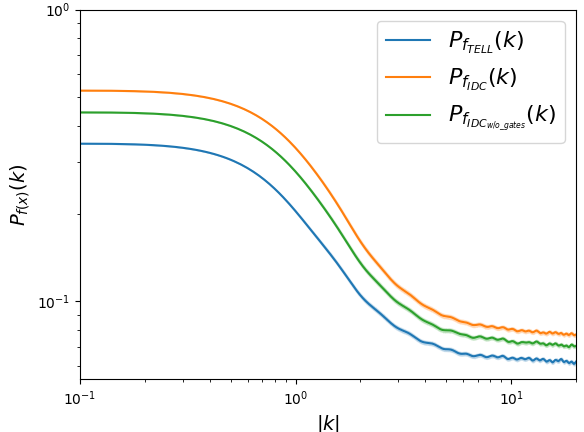}
\caption{Spectral properties of the learned predictive function using \text{ALLAML} dataset. The model trained with the gating network (IDC) has higher Fourier amplitudes at all frequency levels than without gates (IDC$_{w/o\_gates}$) the baseline (TELL). This suggests that IDC can better handle the inductive bias of tabular data.} 
\label{fig:inductive}
\end{center}
\end{figure}

In this section, we analyze whether our local gates, which sparsify the weights of our learned prediction function, can provide an inductive bias that helps the network learn high-frequency functions. To test this, we trained our model on the tabular ALLAML data and evaluated the absolute value of Fourier amplitudes of predictions of our model $|\text{NUDFT}(f_{\text{IDC}})|$ at different frequencies $|k|$. In Figure \ref{fig:inductive}, we compare these amplitudes to those learned without the gates (IDC$_{\text{w/o\_gates}}$) and to those learned by the interpretable deep clustering model TELL \cite{peng2022xai}. This figure demonstrates that the sparsification helps the model learn high-frequency components. The bias induced by the gating network makes our model more accurate on tabular data while being interpretable.

\subsection{Ablation Study}
\label{sec:ablation}
We conducted an ablation study to determine if all the components of the  loss are necessary for the model to converge. The evaluation was carried out on the $\text{MNIST}_{60K}$ dataset, and we ran each experiment 10 times. The results are presented in Table \ref{tab:ablation}. We train our model without regularization term in Eq.\ref{eq:reg} (row 2), without latent or input augmentations (rows 3-4), then we replace the clustering head with $K$-Means (row 5) and finally train without reconstruction term $\mathcal{L}_{\text{recon}}$. Our findings indicate that all the proposed components significantly contribute to the performance of the model. 

\section{Conclusions}

We present a deep clustering model that accurately assigns clusters on tabular data and predicts informative features at both the sample and cluster levels, providing interpretability. We tested our model on 15 datasets including synthetic datasets, high-dimensional tabular datasets, and several image datasets (treated as tables). A main limitation of our scheme is dealing with correlated variables. This is a known caveat of the reconstruction loss \citep{abid2019concrete}.  One way to alleviate this limitation is to incorporate a group sparsity loss as presented by \citep{imrie2022composite}. Our method also struggles to handle datasets with a large number of clusters. This problem could be mitigated using weak supervision or by introducing several clustering heads to prevent collapse to a subset of clusters. We hope our work will be beneficial to scientists in the biomedical field.

\section*{Impact Statement}
The proposed framework for deep learning clustering in general domain tabular data has wide applications in scientific research, healthcare, and biomedicine. It addresses a critical need in data analysis across diverse domains by delivering reliable and interpretable cluster assignments. Biologists can use the framework to analyze genome sequences, medical records, and images with precision. Our hope is that it will increase trust in deep learning-based clustering models for multidisciplinary datasets, such as biological, text, image, and physics tabular datasets. Nonetheless, the framework can suffer from fairness biases at a societal level since it is fully unsupervised and could reflect data biases. We note that this concern is not different from any other unsupervised model, and we leave this as an open issue for future work. 

\section*{Acknowledgements}

We thank the anonymous reviewers for constructive comments during the reviewing process. In addition, we would like to express our deepest gratitude to Uri Shaham (Bar Ilan University) for insightful and constructive discussions significantly enriched this research. 


\bibliography{references}
\bibliographystyle{icml2024}

\newpage
\appendix
\onecolumn
\input{supp}

\end{document}

%% file: math_commands.tex
\usepackage{amsmath,amsfonts,bm}

\def\eqref#1{equation~\ref{#1}}

\def\1{\bm{1}}

\DeclareMathAlphabet{\mathsfit}{\encodingdefault}{\sfdefault}{m}{sl}
\SetMathAlphabet{\mathsfit}{bold}{\encodingdefault}{\sfdefault}{bx}{n}

\def\sR{{\mathbb{R}}}



%% file: supp.tex
\begin{appendix}

\section{Implementation details}

We implement our model in Pytorch and run experiments on Nvidia A100 GPU server with Intel(R) Xeon(R) Gold 6338 CPU @ 2.00GHz.

\section{Model Architecture}

The model is trained with a single hidden layer in the Clustering Head and the Gating NN. We use up to 4 hidden layers for the Encoder, and The Decoder is a mirrored version of the encoder. The dimensions of hidden layers are detailed in the supplementary included code and will be released to GitHub.

\section{Training Setup}

We train all models with a two-stage approach - we train Encoder, Decoder, and Gating NN in the first stage and then train Clustering Head in the second stage. 
For interpretabiltiy experiments, we train $K$-means \footnote{https://scikit-learn.org/stable/modules/generated/sklearn.cluster.KMeans.html} and TELL \citep{peng2022xai} \footnote{The implementation was found here: https://github.com/XLearning-SCU/2022-JMLR-TELL/tree/main, accessed on 2023-09-28.}. Both methods are trained without additional augmentations for fair comparison to our method with the provided default experimental parameters. 
For small-sample datasets we increase the number of epochs in TELL model up to 10K for fair comparison. In method k-DVAE \footnote{https://github.com/aviclu/k-DVAE} \cite{caciularu2022entangledmixture} is trained with default parameters provided in the shared code with only incresed number of epochs to 200 (initial step) and to 100 (other steps) for small-sample datasets.

\begin{table}[h]
\vskip -0.2in
\centering
\caption{The number of epochs and batch size for different datasets.}
\scalebox{0.8}{
\begin{tabular}{|l |c |c| l |}
 \hline
Dataset   & Epochs Stage 1 & Epochs Stage 2 & Batch size \\ [0.5ex] \hline
Synthetic      & 50 & 2000 & 800  \\ 
$\text{MNIST}_{60K}$           & 300 & 600 & 256 \\ 
$\text{MNIST}_{10K}$           & 300 & 700 & 100 \\ 
FashionMNIST     & 100 & 500 & 256 \\ 
TOX-171   & 1000 & 1000 & 16 \\ 
ALLAML     & 1000 & 1000 & 36 \\ 
PROSTATE    & 1000 & 1000 & 102 \\ 
SRBCT     & 2000 & 1000 & 83 \\ 
BIASE       & 10000 & 1000 & 56 \\ 
INTESTINE      & 5000 & 1000 & 238 \\ 
PBMC-2      & 100 & 100 & 256  \\ 
CNAE-9 & 1000 & 1000 & 500 \\
MFEATZ & 1000 & 1000 & 500 \\
MiniBooNE & 20 & 30 & 512 \\
ALBERT & 10 & 40 & 1024 \\
CIFAR-10 & 600 & 700 & 256 \\
\hline
\end{tabular}
}
\label{tab:technical}
\vskip -0.3in
\end{table}

\section{Regularization Term}
\begin{figure}[ht]
\begin{center} 
\vskip -0.2in
\includegraphics[width=0.5\textwidth]{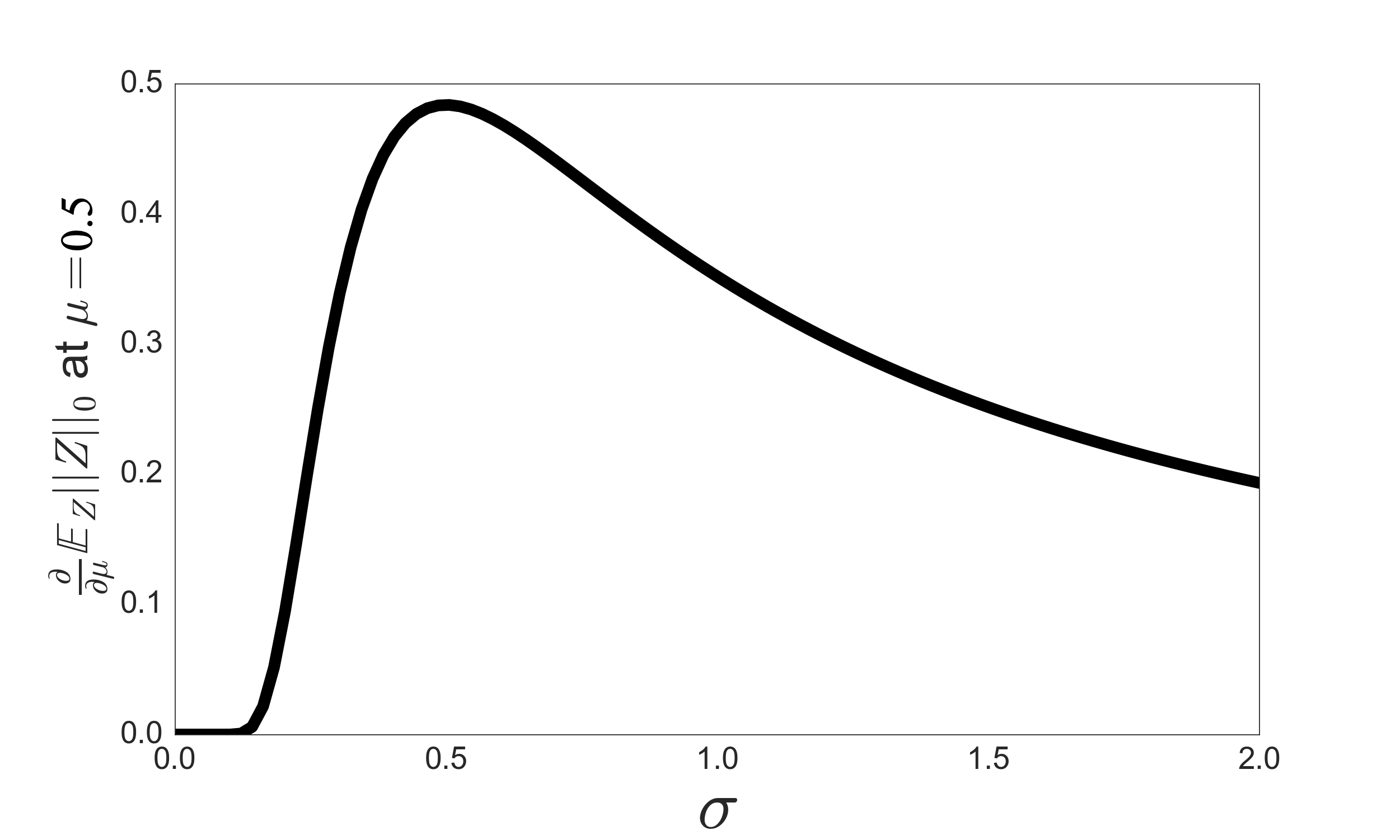}
\end{center}
\vskip -0.1in
\caption{The value of $\frac{\partial}{\partial \mu}\mathbb{E}_Z ||\myvec{Z}||_0 \vert _{\mu=0.5} = \frac{1}{\sqrt{2\pi \sigma^2}} e^{-{\frac{1}{8\sigma^2}}}$ for $\sigma = [0.001, 2]$.}
\label{fig:reg_grad}
\end{figure}
The leading term in our regularizer is expressed by : 

\begin{align*}
    \mathbb{E}_Z ||\myvec{Z}||_0 &= \sum_{d=1}^D \mathbb{P}[z_d > 0] = \sum_{d=1}^D \mathbb{P}[\mu_d + \sigma \epsilon_d +0.5> 0] \\
    &= \sum_{d=1}^D \{ 1 - \mathbb{P}[\mu_d + \sigma \epsilon_d +0.5 \le 0] \} \\
    &= \sum_{d=1}^D \{ 1 - \Phi(\frac{-\mu_d-0.5}{\sigma}) \} \\
    &= \sum_{d=1}^D \Phi\left (\frac{\mu_d+0.5}{\sigma} \right) \\
    &= \sum^D_{d=1} \left( \frac{1}{2} - \frac{1}{2} \erf \left(-\frac{\mu_d+0.5}{\sqrt{2}\sigma} \right) \right)
\end{align*}

To tune $\sigma$, we follow the suggestion in \cite{yamada2020feature}. Specifically, the effect of $\sigma$ can be understood by looking at the value of $\frac{\partial}{\partial \mu_d} \mathbb{E}_Z || \myvec{Z}||_0$. 
In the first training step, $\mu_d$ is $0$. Therefore, at initial training phase, $\frac{\partial}{\partial \mu_d} \mathbb{E}_Z || \myvec{Z}||_0$ is close to $\frac{1}{\sqrt{2\pi \sigma_d^2}} e^{-{\frac{1}{8\sigma_d^2}}}$. To enable sparsification, this term (multiplied by the regularization parameter $\lambda$) has to be greater than the derivative of the loss with respect to $\mu_d$ because otherwise $\mu_d$ is updated in the incorrect direction.
To encourage such behavior, we tune $\sigma$ to the value that maximizes the gradient of the regularization term. As demonstrated in Fig. \ref{fig:reg_grad} this is obtained when $\sigma=0.5$. Therefore, we keep $\sigma=0.5$ throughout our experiments unless specifically noted.

\section{Datasets Properties and References}
\label{app:datasets}
In Table \ref{tab:dataref} we add the references of the datasets used in the experiments. We provide here a short description for each dataset:
\begin{itemize}
    \item \textbf{MNIST$_{10K}$ and MNIST$_{60K}$} are the subsets of MNIST \cite{lecunMNISTHandwritten} dataset, the first one of size 10,000 samples and the second one is the full trainset. Additionally, we use the MNIST test set split for evaluations of the model on unseen data. The images include hand-written digits from 10 categories.
    
    \item \textbf{FashionMNIST$_{60K}$} is a train set of \cite{githubGitHubZalandoresearchfashionmnist}. The images include 10 categories of clothers.
    
    \item \textbf{TOX-171} \cite{piloto2010ovo1} dataset is an example of the use of toxicology to integrate diverse biological data, such as clinical chemistry, expression, and other types of data. The database contains the profiles resulting from the three toxicants: alpha-naphthyl-isothiocyanate, dimethylnitrosamine, and N-methylformamide administered to rats. The classification task is to identify whether the samples are toxic, non toxic or control.

    \item \textbf{ALLAML} dataset \cite{golub1999molecular} consists of gene expression profiles of two acute cases of leukemia: acute lymphoblastic leukemia (ALL) and acute myeloblastic leukemia (AML). The ALL part of the dataset comes from two types, B-cell and T-cell, while the AML part is split into two types, bone marrow samples and peripheral blood samples. It contains in total 72 samples in 2 classes, ALL and AML, which have 47 and 25 samples, respectively. Every sample contains 7,129 gene expression values. 

    \item \textbf{PROSTATE} dataset \cite{singh2002gene} has in total 102 samples in two classes tumor and normal, which have 52 and 50 samples, respectively. After preprocessing described in \cite{nie2010efficient}, a data set contains 102 samples and 5966 genes. 
    
    \item \textbf{SRBCT} dataset \cite{kar2015gene}  consists of four classes of cancers in 83 samples. These four classes were ewing sarcoma (EWS), non-Hodgkin lymphoma (NHL), neuroblastoma (NB), and rhabdomyosarcoma (RMS).
    \item \textbf{BIASE} dataset \cite{biase2014cell} is bimodal mRNA expressions to embryonic genome activation and it contains 56 samples each one of dimension 25,683.
    \item INTESTINE \cite{sato2009single} is a mouse intestine dataset with 238 samples.
    \item \textbf{PBMC-2} dataset is a binary-class subset of the original PBMC \cite{zheng2017massively} dataset. We select two categories that have the most number of samples in the original set. In addition, we remove all zero columns from the data resulting in 17,126 featurees $\times$ 20,742 samples size.
    \item \textbf{CIFAR10} dataset  \cite{krizhevsky2009learning} contains 60,000 small images of size $32 \times 32$ with 3 color channels. In total there are 3,072 features.
    \item \textbf{cnae-9} \footnote{\text{https://www.openml.org/search?type=data\&sort=runs\&id=1468}. Accessed on 2024-01-23} \cite{bischl2017openml, ciarelli2009agglomeration} contains 1080 documents of free text business descriptions of Brazilian companies categorized into a subset of 9 categories. The original texts were preprocessed to obtain the current data set: initially, it was kept only letters and then it was removed prepositions of the texts. Next, the words were transformed to their canonical form. Finally, each document was represented as a vector, where the weight of each word is its frequency in the document. This data set is highly sparse (99.22\% of the matrix is filled with zeros).
    \item \textbf{MFEATZ} is known as mfeat-zernike  \footnote{\text{https://www.openml.org/search?type=data\&status=active\&id=22}. Accessed on 2024-01-23} \cite{van1998handwritten, bischl2017openml} dataset describes features of handwritten numerals (0 - 9) extracted from a collection of Dutch utility maps. Corresponding patterns in different datasets correspond to the same original character. 200 instances per class (for a total of 2,000 instances) have been digitized in binary images. These digits are represented in terms of 47 Zernike moments.

    \item \textbf{MiniBooNE} is a physical dataset for particle identification task \cite{roe2005boosted}.

    \item \textbf{ALBERT} is a text dataset from AutoML challenge \cite{automlchallenges}. Since the dataset comes with mnostly categorical variables, we follow the work \cite{beyazit2023inductive} and preprocess the dataset with target-encoder\footnote{https://scikit-learn.org/stable/modules/generated/sklearn.preprocessing.TargetEncoder.html}.
   
\end{itemize}
\begin{table}[ht]
\vskip -0.2in
\label{tab:dataref}
\centering
\caption{Properties and references for the dataset used in the experiments.}
\resizebox{0.7\linewidth}{!}{
\begin{tabular}{|l|c|c|c|c|}
                \hline 
              Dataset & Features & Samples & Clusters & Reference \\
                 \hline 
            MNIST$_{10K}$ & 784 & 10,000 & 10 & \cite{lecunMNISTHandwritten} \\
            MNIST$_{60K}$ & 784 & 60,000 & 10 & \cite{lecunMNISTHandwritten} \\
            FashionMNIST$_{60K}$ & 784 & 60,000 & 10 & \cite{githubGitHubZalandoresearchfashionmnist} \\
            TOX-171 & 5,748 & 171 & 4 & \cite{jundonglDatasetsFeature}\\
            ALLAML & 7,192 & 72 & 2 & \cite{jundonglDatasetsFeature} \\
            PROSTATE & 5,966 & 102 & 2 & \cite{jundonglDatasetsFeature} \\
            SRBCT & 2,308 & 83 & 4 & \cite{khan2001classification} \\
            BIASE & 25,683 & 56 & 4 & \cite{biase2014cell} \\
            INTESTINE & 3,775 & 238 & 13 & \cite{sato2009single} \\
            PBMC-2 & 17,126 & 20,742 & 2 & \cite{zheng2017massively} \\
            CIFAR10 & 3,072 & 60,000 & 10 & \cite{krizhevsky2009learning} \\
            CNAE-9 & 857 & 1080 & 9 & \cite{ciarelli2009agglomeration, bischl2017openml} \\
            MFEATZ & 48 & 2000 &  2 & \cite{van1998handwritten, bischl2017openml} \\
            MiniBooNE & 50 & 130064 & 2 & \cite{roe2005boosted, bischl2017openml}\\
            ALBERT & 78 & 425240 & 2 & \cite{automlchallenges, bischl2017openml}\\ 

        \hline
 \end{tabular}}

\end{table}

\section{Train Loss Augmentations}
\label{app:augmentations}
In addition to the loss presented in Section 4.2 we exploit the next dataset-agnostic augmentations during model training. The first one is the standard reconstruction loss that is calculated between input samples and reconstructed samples. Input denoising is based on \cite{vincent2008extracting} and latent denoising on \cite{doi2007robust}:
\begin{itemize}
    \item  Clean reconstruction loss, $||f_{\theta_D} \circ f_{\theta_E}(\myvec{x}_i) - \myvec{x}_i||_1$, which measures the deviation of estimated $\hat{\myvec{x}_i}$ from the input sample $\myvec{x}_i$.
    
    \item Denoising reconstruction loss \cite{vincent2008extracting}, $||f_{\theta_D} \circ f_{\theta_E} (\myvec{x}_i \odot m_{rand}) - \myvec{x}_i||_1$, where $m_{rand} \in \{0,1\}^{D}$ is a random binary mask generated for each sample $\myvec{x}_i$. We generate a mask such that about $m_{rand} \%$ of the input features are multiplied by zero value, which indicates that the gate is closed. The loss pushes the method to pay less attention to unnecessary features for the reconstruction.
    
    \item Latent denoising reconstruction loss, $||f_{\theta_D}(\myvec{h}_i \odot \myvec{h}_{noise}) - \myvec{x}_i||_1$, where $h_{noise} \sim \mathcal{N}(1, \sigma_h)$ is a noise generated from a normal distribution with mean one and scale $\sigma_h$ which is a dataset-specific hyperparameter \cite{doi2007robust}. This term aims to improve latent embedding representation by small perturbation augmentation to treat small sample-size datasets.

\end{itemize}

\section{Model Training Scalability}
In Figure \ref{plot:trainscale} we show the training time as a function of number of data samples. It could be seen that training time scales linearly with an increase in dataset length. In addition, in Table \ref{table:mnist_subset} we present the total training time (in seconds) measured for the baselines and our method. The experiments were done on a single A100 GPU. The measurements were done for different dataset (MNIST) sizes and averaged over 5 trials with different random seeds (700 epochs, batch size 128, time in seconds).

\begin{table}[h!]
\centering
\begin{minipage}{0.45\linewidth}
\centering
\caption{Training time in seconds for different MNIST subset sizes.}
\resizebox{0.8\linewidth}{!}{
\begin{tabular}{cccc}
\hline
\textbf{MNIST subset} & \textbf{IDC} & \textbf{TELL} & \textbf{k-DVAE} \\ 
\hline
100 & 181.4 & 1.8 & 487.1 \\ 
1000 & 401.3 & 69.3 & 1647.9 \\ 
10000 & 2016.2 & 735.5 & 9705.6 \\ 
\hline
\end{tabular}}

\label{table:mnist_subset}
\end{minipage}
\begin{minipage}{0.45\linewidth}
\centering
\includegraphics[width=\linewidth]{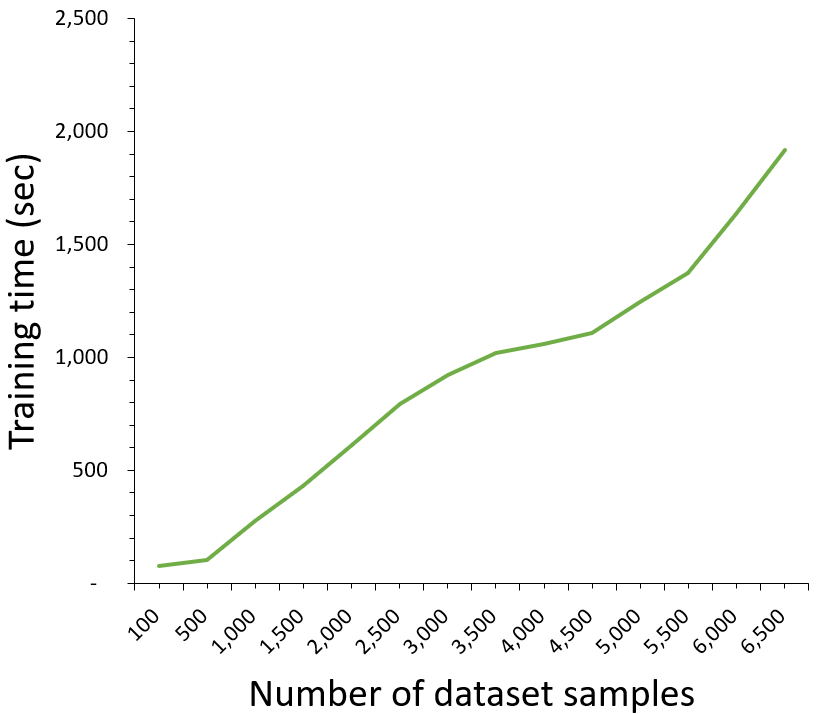}
\vskip -0.1in
\caption{Training time in seconds measured for different numbers of samples.}
\label{plot:trainscale}
\end{minipage}
\vskip -0.3in
\end{table}

\section{Synthetic Dataset Generation}
\label{app:synthgen}

The dataset consists of three informative features $x_i[j] \in [-1,1],j=1,..,3$ for each sample $\myvec{x}_i$ and is generated as isotropic Gaussian blobs \footnote{\text{https://scikit-learn.org/stable/modules/generated/sklearn.datasets.make\_blobs.html}} with standard deviation of each cluster of $0.5$.  The detailed description of the dataset generation could be found in \ref{app:synthgen}. 
Then we add ten nuisance background features with values drawn from $\mathcal{N}(0, \sigma_n^2)$ (with $\sigma_n=0.1$) resulting in 13 total features. The samples are equally distributed between 4 clusters, with $\sim800$ samples in each cluster. Given the first two dimensions $\{x[1], x[2]\}$, only 3 clusters are separable, and the same property holds for dimensions pair $\{x[1], x[3]\}$.

\section{Indactive Bias Analysis}
\label{app:inductive}
We present the original plot from the paper \cite{beyazit2023inductive} that emphasizes the difference in target labels Fourier amplitudes distribution $P_y$ across different frequencies. The authors claim that distributiuon of Foueirer amplitueds obtaned on ground truth targets on tabular datasets has higher values than those of image datasets. 
We support this claim by testing it from the learned function - does our model learns a bias for tabular domain by using gating network. To produce the anaylsis plot we predict 
$\hat{y}_i^j=argmax_{j \in [1,...,K]}(\hat{\myvec{p}}_i)$ where $\hat{\myvec{p}}_i = f_{\theta_C} (f_{\theta_E}(f_G(\myvec{x}_i)))$ and is calculated for each sample $\myvec{x}_i$ and for each feature value $x_i^d \in \myvec{x}_i$. Then we calculate the non uniform discrete fourier transform (NUDFT) for 1000 frequencies values in range [0.1, 20] which accepts a vector of $N$ values of $x_i^d$, $i=1,...N$ and corresponding binary prediction values $y_i$. NUDFT is calculated by the pytorch code:

\begin{lstlisting}[language=Python, basicstyle=\small]]
def spectrum_NUDFT(x, y, kmax=20, nk=1000):
    kvals = np.linspace(0.1, kmax, nk+1)
    nufft = (1 / len(x)) * nfft_adjoint(-(x * kmax / nk), y, 2 * (nk + 1))[nk + 1:]
    return [kvals, np.array(nufft, dtype="complex_")]
\end{lstlisting}

where nfft\_adjoint function is part of the \textbf{nfft} package \footnote{https://github.com/jakevdp/nfft}. Then we take the absolute value of NUDFT amplitudes. By repeating the process for each feature $d$ we obtain $D$ vectors with $|NUDFT|$ values and we plot them with 1000 frequency steps $|\textbf{k}|$ in logarithmic scale axes. In this way we obtain the plot in Figure \ref{fig:inductive}.

\begin{figure}[h]
\label{plot:inductive_Beyazit}
\centering
\includegraphics[trim={0.1cm 0.1cm 0.1cm 0.1cm},clip,width=.45\columnwidth]{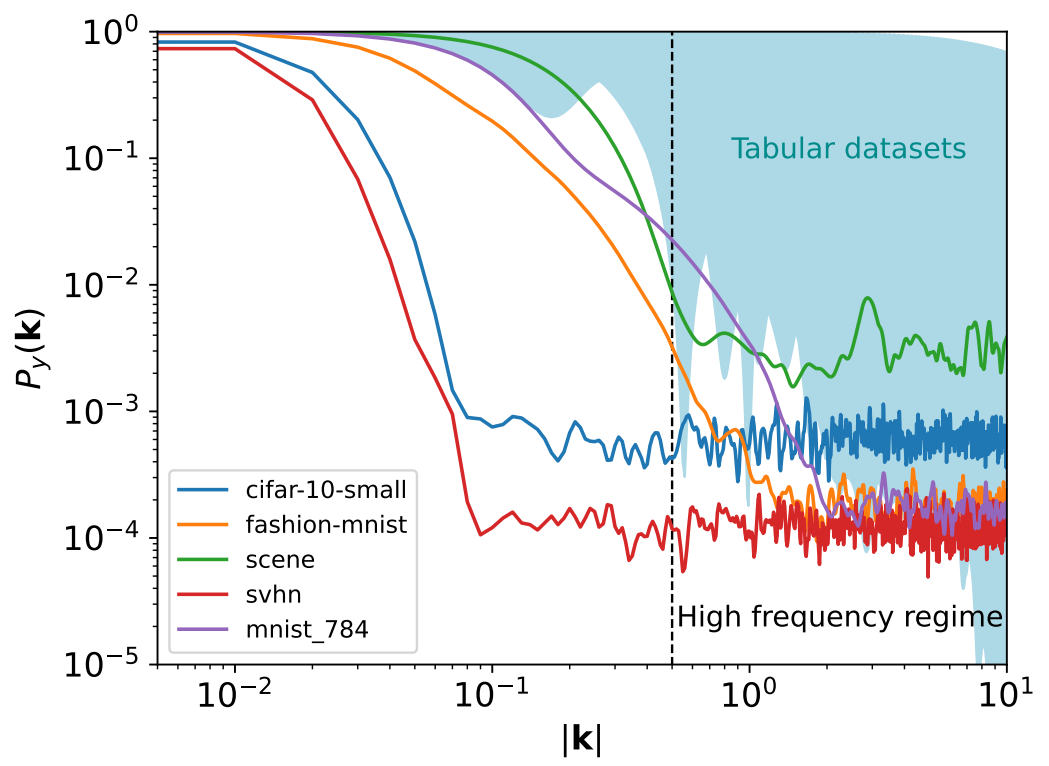}
\vskip -0.1in
\caption{\cite{beyazit2023inductive}: Due to their heterogeneous nature, tabular datasets tend to describe higher frequency target functions compared to images. The spectra corresponding to image datasets (curves in color) tend to feature lower Fourier amplitudes at higher frequencies than hetergoneous tabular datasets (cyan region).}
\vskip -0.2in
\end{figure}

\section{The number of selected features}

\begin{table}[h!]
\centering
\caption{Selected and total features for various datasets}
\resizebox{0.99\linewidth}{!}{
\begin{tabular}{lccccccccccc}
\hline
\textbf{Dataset} & \textbf{TOX-171} & \textbf{ALLAML} & \textbf{PROSTATE} & \textbf{SRBCT} & \textbf{BIASE} & \textbf{INTESTINE} & \textbf{PBMC-2} & \textbf{CNAE-9} & \textbf{MFEATZ} & \textbf{ALBERT} & \textbf{MiniBoonE} \\ 
\hline
\textbf{Selected} & 49.6 & 527.6 & 170.9 & 46.7 & 210 & 65 & 137.8 & 360.5 & 26.2 & 59.4 & 29.4 \\ 
\textbf{Total} & 5748 & 7192 & 5966 & 2308 & 25683 & 3775 & 17126 & 856 & 47 & 78 & 50 \\ 
\hline
\end{tabular}
}
\label{table:datasets}
\vskip -0.1in
\end{table}

\section{Sensitivity analysis of hyper parameters}

 We present the sensitivity analysis of the clustering as a function of hyperparameter changes. The analysis is done on MNIST subset (10K samples), we present the mean values obtained from 5 trials for each parameter change. Please, note that in same cases we get even better results than were presented in the main part of this paper:

\begin{table}[h]
\centering
\vskip -0.1in
\begin{minipage}{0.3\linewidth}
\centering
\caption{Sensitivity to $\epsilon_{gtcr}$ values}
\resizebox{\linewidth}{!}{
\begin{tabular}{ccccc}
\hline
$\epsilon_{gtcr}$ & \textbf{ACC} & \textbf{ARI} & \textbf{NMI} & \textbf{\# open gates} \\ 
\hline
100 & 0.8318 & 0.7533 & 0.7856 & 16 \\ 
10 & 0.7732 & 0.6897 & 0.7386 & 17 \\ 
1 & 0.834 & 0.7479 & 0.7807 & 18 \\ 
0.1 & 0.8067 & 0.7251 & 0.7641 & 18 \\ 
0.01 & 0.8081 & 0.7256 & 0.7667 & 18 \\ 
\hline
\end{tabular}}

\label{table:epsilon_gtcr}
\end{minipage}
\hspace{0.3in}
\begin{minipage}{0.3\linewidth}
\centering
\caption{Sensitivity to $\sigma_h$ values}
\resizebox{\linewidth}{!}{
\begin{tabular}{ccccc}
\hline
$\sigma_h$ & \textbf{ACC} & \textbf{ARI} & \textbf{NMI} & \textbf{\# open gates} \\ 
\hline
0.01 & 0.8104 & 0.7249 & 0.7683 & 18 \\ 
0.1 & 0.8183 & 0.7377 & 0.7736 & 18 \\ 
0.2 & \textbf{0.8531} & \textbf{0.7646} & \textbf{0.7841} & 18 \\ 
0.5 & 0.6158 & 0.4707 & 0.5661 & 18 \\ 
1 & 0.7336 & 0.6115 & 0.6745 & 18 \\ 
\hline
\end{tabular}}
\end{minipage}
\end{table}
\begin{table}[h]
\centering
\begin{minipage}{0.3\linewidth}
\centering
\caption{Sensitivity to $\epsilon_{head}$ values}
\resizebox{\linewidth}{!}{
\begin{tabular}{ccccc}
\hline
$\epsilon_{head}$ & \textbf{ACC} & \textbf{ARI} & \textbf{NMI} & \textbf{\# open gates} \\ 
\hline
100 & 0.6645 & 0.5339 & 0.6294 & 18 \\ 
10 & 0.6025 & 0.503 & 0.6206 & 18 \\ 
1 & 0.7941 & 0.6842 & 0.7298 & 18 \\ 
0.1 & 0.834 & 0.7479 & 0.7807 & 18 \\ 
0.01 & 0.708 & 0.6123 & 0.7122 & 18 \\ 
\hline
\end{tabular}}
\label{table:performance_metrics}
\end{minipage}
\hspace{0.1in}
\begin{minipage}{0.3\linewidth}
\centering
\caption{Sensitivity to $\lambda_{reg}$ values}
\resizebox{\linewidth}{!}{
\begin{tabular}{ccccc}
\hline
$\lambda_{reg}$ & \textbf{ACC} & \textbf{ARI} & \textbf{NMI} & \textbf{\# open gates} \\ 
\hline
0 & \textbf{0.8643} & \textbf{0.7803} & \textbf{0.8021} & 435 \\ 
0.1 & 0.8227 & 0.7471 & 0.7804 & 111 \\ 
1 & 0.834 & 0.7479 & 0.7807 & 18 \\ 
10 & 0.7795 & 0.6921 & 0.7421 & 4 \\ 
100 & 0.1374 & 0.0028 & 0.053 & 0 \\ 
\hline
\end{tabular}}

\label{table:lambda_reg}
\vskip -0.5in
\end{minipage}
\hspace{0.1in}
\begin{minipage}{0.3\linewidth}
\caption{Sensitivity to $m_{rand}$ values}
\resizebox{1\linewidth}{!}{
\begin{tabular}{ccccc}
\hline
$m_{rand}$ & \textbf{ACC} & \textbf{ARI} & \textbf{NMI} & \textbf{\# open gates} \\ 
\hline
0\% & 0.7969 & 0.7131 & 0.7585 & 21 \\ 
10\% & 0.8381 & 0.7539 & 0.7864 & 20 \\ 
30\% & \textbf{0.8812} & \textbf{0.7915} & \textbf{0.8049} & 19 \\ 
50\% & 0.8334 & 0.7491 & 0.7844 & 19 \\ 
70\% & 0.8188 & 0.7464 & 0.7864 & 18 \\ 
90\% & 0.8104 & 0.7249 & 0.7683 & 18 \\ 
\hline
\end{tabular}}

\label{table:masked_features}
\end{minipage}
\end{table}

\end{appendix}

%% file: arxiv.bbl
\begin{thebibliography}{74}
\providecommand{\natexlab}[1]{#1}
\providecommand{\url}[1]{\texttt{#1}}
\expandafter\ifx\csname urlstyle\endcsname\relax
  \providecommand{\doi}[1]{doi: #1}\else
  \providecommand{\doi}{doi: \begingroup \urlstyle{rm}\Url}\fi

\bibitem[git()]{githubGitHubZalandoresearchfashionmnist}
{G}it{H}ub - zalandoresearch/fashion-mnist: {A} {M}{N}{I}{S}{T}-like fashion
  product database. {B}enchmark --- github.com.
\newblock \url{https://github.com/zalandoresearch/fashion-mnist}.

\bibitem[jun()]{jundonglDatasetsFeature}
{D}atasets | {F}eature {S}election @ {A}{S}{U} --- jundongl.github.io.
\newblock \url{https://jundongl.github.io/scikit-feature/datasets.html}.

\bibitem[lec()]{lecunMNISTHandwritten}
{M}{N}{I}{S}{T} handwritten digit database, {Y}ann {L}e{C}un, {C}orinna
  {C}ortes and {C}hris {B}urges --- yann.lecun.com.
\newblock \url{http://yann.lecun.com/exdb/mnist/ }.

\bibitem[Abid et~al.(2019)Abid, Balin, and Zou]{abid2019concrete}
Abid, A., Balin, M.~F., and Zou, J.
\newblock Concrete autoencoders for differentiable feature selection and
  reconstruction.
\newblock \emph{arXiv preprint arXiv:1901.09346}, 2019.

\bibitem[Alvarez~Melis \& Jaakkola(2018)Alvarez~Melis and
  Jaakkola]{alvarez2018towards}
Alvarez~Melis, D. and Jaakkola, T.
\newblock Towards robust interpretability with self-explaining neural networks.
\newblock \emph{Advances in neural information processing systems}, 31, 2018.

\bibitem[Armanfard et~al.(2015)Armanfard, Reilly, and
  Komeili]{armanfard2015local}
Armanfard, N., Reilly, J.~P., and Komeili, M.
\newblock Local feature selection for data classification.
\newblock \emph{IEEE transactions on pattern analysis and machine
  intelligence}, 38\penalty0 (6):\penalty0 1217--1227, 2015.

\bibitem[Armingol et~al.(2021)Armingol, Officer, Harismendy, and
  Lewis]{armingol2021deciphering}
Armingol, E., Officer, A., Harismendy, O., and Lewis, N.~E.
\newblock Deciphering cell--cell interactions and communication from gene
  expression.
\newblock \emph{Nature Reviews Genetics}, 22\penalty0 (2):\penalty0 71--88,
  2021.

\bibitem[Bal{\i}n et~al.(2019)Bal{\i}n, Abid, and Zou]{balin2019concrete}
Bal{\i}n, M.~F., Abid, A., and Zou, J.
\newblock Concrete autoencoders: Differentiable feature selection and
  reconstruction.
\newblock In \emph{International conference on machine learning}, pp.\
  444--453. PMLR, 2019.

\bibitem[Basri et~al.(2020)Basri, Galun, Geifman, Jacobs, Kasten, and
  Kritchman]{basri2020frequency}
Basri, R., Galun, M., Geifman, A., Jacobs, D., Kasten, Y., and Kritchman, S.
\newblock Frequency bias in neural networks for input of non-uniform density.
\newblock In \emph{International Conference on Machine Learning}, pp.\
  685--694. PMLR, 2020.

\bibitem[Bertsimas et~al.(2021)Bertsimas, Orfanoudaki, and
  Wiberg]{bertsimas2021interpretable}
Bertsimas, D., Orfanoudaki, A., and Wiberg, H.
\newblock Interpretable clustering: an optimization approach.
\newblock \emph{Machine Learning}, 110:\penalty0 89--138, 2021.

\bibitem[Beyazit et~al.(2023)Beyazit, Kozaczuk, Li, Wallace, and
  Fadlallah]{beyazit2023inductive}
Beyazit, E., Kozaczuk, J., Li, B., Wallace, V., and Fadlallah, B.~H.
\newblock An inductive bias for tabular deep learning.
\newblock In \emph{Thirty-seventh Conference on Neural Information Processing
  Systems}, 2023.

\bibitem[Biase et~al.(2014)Biase, Cao, and Zhong]{biase2014cell}
Biase, F.~H., Cao, X., and Zhong, S.
\newblock Cell fate inclination within 2-cell and 4-cell mouse embryos revealed
  by single-cell rna sequencing.
\newblock \emph{Genome research}, 24\penalty0 (11):\penalty0 1787--1796, 2014.

\bibitem[Bischl et~al.(2017)Bischl, Casalicchio, Feurer, Gijsbers, Hutter,
  Lang, Mantovani, van Rijn, and Vanschoren]{bischl2017openml}
Bischl, B., Casalicchio, G., Feurer, M., Gijsbers, P., Hutter, F., Lang, M.,
  Mantovani, R.~G., van Rijn, J.~N., and Vanschoren, J.
\newblock Openml benchmarking suites.
\newblock \emph{arXiv preprint arXiv:1708.03731}, 2017.

\bibitem[Bregman et~al.(2021)Bregman, Lindenbaum, and Rabin]{bregman2021array}
Bregman, Y., Lindenbaum, O., and Rabin, N.
\newblock Array based earthquakes-explosion discrimination using diffusion
  maps.
\newblock \emph{Pure and Applied Geophysics}, 178:\penalty0 2403--2418, 2021.

\bibitem[Caciularu \& Goldberger(2023)Caciularu and
  Goldberger]{caciularu2022entangledmixture}
Caciularu, A. and Goldberger, J.
\newblock An entangled mixture of variational autoencoders approach to deep
  clustering.
\newblock \emph{Neurocomputing}, 2023.

\bibitem[Cai et~al.(2022)Cai, Fan, Guo, Wang, Zhang, and
  Zhang]{cai2022efficient}
Cai, J., Fan, J., Guo, W., Wang, S., Zhang, Y., and Zhang, Z.
\newblock Efficient deep embedded subspace clustering.
\newblock In \emph{Proceedings of the IEEE/CVF Conference on Computer Vision
  and Pattern Recognition}, pp.\  1--10, 2022.

\bibitem[Ciarelli \& Oliveira(2009)Ciarelli and
  Oliveira]{ciarelli2009agglomeration}
Ciarelli, P.~M. and Oliveira, E.
\newblock Agglomeration and elimination of terms for dimensionality reduction.
\newblock In \emph{2009 Ninth International Conference on Intelligent Systems
  Design and Applications}, pp.\  547--552. IEEE, 2009.

\bibitem[Cohen(2023)]{cohen2023interpretable}
Cohen, E.
\newblock Interpretable clustering via soft clustering trees.
\newblock In \emph{International Conference on Integration of Constraint
  Programming, Artificial Intelligence, and Operations Research}, pp.\
  281--298. Springer, 2023.

\bibitem[Deprez et~al.(2020)Deprez, Zaragosi, Truchi, Becavin,
  Ruiz~Garc{\'\i}a, Arguel, Plaisant, Magnone, Lebrigand, Abelanet,
  et~al.]{deprez2020single}
Deprez, M., Zaragosi, L.-E., Truchi, M., Becavin, C., Ruiz~Garc{\'\i}a, S.,
  Arguel, M.-J., Plaisant, M., Magnone, V., Lebrigand, K., Abelanet, S., et~al.
\newblock A single-cell atlas of the human healthy airways.
\newblock \emph{American journal of respiratory and critical care medicine},
  202\penalty0 (12):\penalty0 1636--1645, 2020.

\bibitem[Doi et~al.(2007)Doi, Balcan, and Lewicki]{doi2007robust}
Doi, E., Balcan, D.~C., and Lewicki, M.~S.
\newblock Robust coding over noisy overcomplete channels.
\newblock \emph{IEEE Transactions on Image Processing}, 16\penalty0
  (2):\penalty0 442--452, 2007.

\bibitem[Frost et~al.(2020)Frost, Moshkovitz, and Rashtchian]{frost2020exkmc}
Frost, N., Moshkovitz, M., and Rashtchian, C.
\newblock Exkmc: Expanding explainable $ k $-means clustering.
\newblock \emph{arXiv preprint arXiv:2006.02399}, 2020.

\bibitem[Gabidolla \& Carreira-Perpi{\~n}{\'a}n(2022)Gabidolla and
  Carreira-Perpi{\~n}{\'a}n]{gabidolla2022optimal}
Gabidolla, M. and Carreira-Perpi{\~n}{\'a}n, M.~{\'A}.
\newblock Optimal interpretable clustering using oblique decision trees.
\newblock In \emph{Proceedings of the 28th ACM SIGKDD Conference on Knowledge
  Discovery and Data Mining}, pp.\  400--410, 2022.

\bibitem[Gao et~al.(2020)Gao, Yang, Gouk, and Hospedales]{gao2020deep}
Gao, B., Yang, Y., Gouk, H., and Hospedales, T.~M.
\newblock Deep clusteringwith concrete k-means.
\newblock In \emph{Icassp 2020-2020 ieee international conference on acoustics,
  speech and signal processing (icassp)}, pp.\  4252--4256. IEEE, 2020.

\bibitem[Golub et~al.(1999)Golub, Slonim, Tamayo, Huard, Gaasenbeek, Mesirov,
  Coller, Loh, Downing, Caligiuri, et~al.]{golub1999molecular}
Golub, T.~R., Slonim, D.~K., Tamayo, P., Huard, C., Gaasenbeek, M., Mesirov,
  J.~P., Coller, H., Loh, M.~L., Downing, J.~R., Caligiuri, M.~A., et~al.
\newblock Molecular classification of cancer: class discovery and class
  prediction by gene expression monitoring.
\newblock \emph{science}, 286\penalty0 (5439):\penalty0 531--537, 1999.

\bibitem[Guan et~al.(2011)Guan, Jordan, and Dy]{guan2011unified}
Guan, Y., Jordan, M.~I., and Dy, J.~G.
\newblock A unified probabilistic model for global and local unsupervised
  feature selection.
\newblock In \emph{Proceedings of the 28th International Conference on Machine
  Learning (ICML-11)}, pp.\  1073--1080, 2011.

\bibitem[Guyon et~al.(2019)Guyon, Sun-Hosoya, Boull\'e, Escalante, Escalera,
  Liu, Jajetic, Ray, Saeed, Sebag, Statnikov, Tu, and Viegas]{automlchallenges}
Guyon, I., Sun-Hosoya, L., Boull\'e, M., Escalante, H.~J., Escalera, S., Liu,
  Z., Jajetic, D., Ray, B., Saeed, M., Sebag, M., Statnikov, A., Tu, W., and
  Viegas, E.
\newblock Analysis of the automl challenge series 2015-2018.
\newblock In \emph{AutoML}, Springer series on Challenges in Machine Learning,
  2019.
\newblock URL
  \url{https://www.automl.org/wp-content/uploads/2018/09/chapter10-challenge.pdf}.

\bibitem[Han et~al.(2018)Han, Wang, Zhang, Li, and Xu]{han2018autoencoder}
Han, K., Wang, Y., Zhang, C., Li, C., and Xu, C.
\newblock Autoencoder inspired unsupervised feature selection.
\newblock In \emph{2018 IEEE international conference on acoustics, speech and
  signal processing (ICASSP)}, pp.\  2941--2945. IEEE, 2018.

\bibitem[He et~al.(2005)He, Cai, and Niyogi]{he2005laplacian}
He, X., Cai, D., and Niyogi, P.
\newblock Laplacian score for feature selection.
\newblock \emph{Advances in neural information processing systems}, 18, 2005.

\bibitem[Huang et~al.(2019)Huang, Cai, and Wang]{huang2019unsupervised}
Huang, D., Cai, X., and Wang, C.-D.
\newblock Unsupervised feature selection with multi-subspace randomization and
  collaboration.
\newblock \emph{Knowledge-Based Systems}, 182:\penalty0 104856, 2019.

\bibitem[Imrie et~al.(2022)Imrie, Norcliffe, Li{\`o}, and van~der
  Schaar]{imrie2022composite}
Imrie, F., Norcliffe, A., Li{\`o}, P., and van~der Schaar, M.
\newblock Composite feature selection using deep ensembles.
\newblock \emph{Advances in Neural Information Processing Systems},
  35:\penalty0 36142--36160, 2022.

\bibitem[Jana et~al.(2023)Jana, Li, Yamada, and Lindenbaum]{jana2023support}
Jana, S., Li, H., Yamada, Y., and Lindenbaum, O.
\newblock Support recovery with projected stochastic gates: Theory and
  application for linear models.
\newblock \emph{Signal Processing}, 213:\penalty0 109193, 2023.

\bibitem[Jang et~al.(2016)Jang, Gu, and Poole]{jang2016categorical}
Jang, E., Gu, S., and Poole, B.
\newblock Categorical reparameterization with gumbel-softmax.
\newblock \emph{arXiv preprint arXiv:1611.01144}, 2016.

\bibitem[Jiang et~al.(2016)Jiang, Zheng, Tan, Tang, and
  Zhou]{jiang2016variational}
Jiang, Z., Zheng, Y., Tan, H., Tang, B., and Zhou, H.
\newblock Variational deep embedding: An unsupervised and generative approach
  to clustering.
\newblock \emph{arXiv preprint arXiv:1611.05148}, 2016.

\bibitem[Kar et~al.(2015)Kar, Sharma, and Maitra]{kar2015gene}
Kar, S., Sharma, K.~D., and Maitra, M.
\newblock Gene selection from microarray gene expression data for
  classification of cancer subgroups employing pso and adaptive k-nearest
  neighborhood technique.
\newblock \emph{Expert Systems with Applications}, 42\penalty0 (1):\penalty0
  612--627, 2015.

\bibitem[Kauffmann et~al.(2022)Kauffmann, Esders, Ruff, Montavon, Samek, and
  M{\"u}ller]{kauffmann2022clustering}
Kauffmann, J., Esders, M., Ruff, L., Montavon, G., Samek, W., and M{\"u}ller,
  K.-R.
\newblock From clustering to cluster explanations via neural networks.
\newblock \emph{IEEE Transactions on Neural Networks and Learning Systems},
  2022.

\bibitem[Khan et~al.(2001)Khan, Wei, Ringner, Saal, Ladanyi, Westermann,
  Berthold, Schwab, Antonescu, Peterson, et~al.]{khan2001classification}
Khan, J., Wei, J.~S., Ringner, M., Saal, L.~H., Ladanyi, M., Westermann, F.,
  Berthold, F., Schwab, M., Antonescu, C.~R., Peterson, C., et~al.
\newblock Classification and diagnostic prediction of cancers using gene
  expression profiling and artificial neural networks.
\newblock \emph{Nature medicine}, 7\penalty0 (6):\penalty0 673--679, 2001.

\bibitem[Kiselev et~al.(2017)Kiselev, Kirschner, Schaub, Andrews, Yiu, Chandra,
  Natarajan, Reik, Barahona, Green, et~al.]{kiselev2017sc3}
Kiselev, V.~Y., Kirschner, K., Schaub, M.~T., Andrews, T., Yiu, A., Chandra,
  T., Natarajan, K.~N., Reik, W., Barahona, M., Green, A.~R., et~al.
\newblock Sc3: consensus clustering of single-cell rna-seq data.
\newblock \emph{Nature methods}, 14\penalty0 (5):\penalty0 483--486, 2017.

\bibitem[Krizhevsky et~al.(2009)Krizhevsky, Hinton,
  et~al.]{krizhevsky2009learning}
Krizhevsky, A., Hinton, G., et~al.
\newblock Learning multiple layers of features from tiny images.
\newblock 2009.

\bibitem[Lawless et~al.(2022)Lawless, Kalagnanam, Nguyen, Phan, and
  Reddy]{lawless2022interpretable}
Lawless, C., Kalagnanam, J., Nguyen, L.~M., Phan, D., and Reddy, C.
\newblock Interpretable clustering via multi-polytope machines.
\newblock In \emph{Proceedings of the aaai conference on artificial
  intelligence}, volume~36, pp.\  7309--7316, 2022.

\bibitem[Lee et~al.(2022)Lee, Imrie, and van~der Schaar]{lee2022self}
Lee, C., Imrie, F., and van~der Schaar, M.
\newblock Self-supervision enhanced feature selection with correlated gates.
\newblock In \emph{International Conference on Learning Representations}, 2022.

\bibitem[Li et~al.(2022)Li, Chen, LeCun, and Sommer]{li2022neural}
Li, Z., Chen, Y., LeCun, Y., and Sommer, F.~T.
\newblock Neural manifold clustering and embedding.
\newblock \emph{arXiv preprint arXiv:2201.10000}, 2022.

\bibitem[Lindenbaum et~al.(2021{\natexlab{a}})Lindenbaum, Nouri, Kluger, and
  Kleinstein]{lindenbaum2021alignment}
Lindenbaum, O., Nouri, N., Kluger, Y., and Kleinstein, S.~H.
\newblock Alignment free identification of clones in b cell receptor
  repertoires.
\newblock \emph{Nucleic acids research}, 49\penalty0 (4):\penalty0 e21--e21,
  2021{\natexlab{a}}.

\bibitem[Lindenbaum et~al.(2021{\natexlab{b}})Lindenbaum, Shaham, Peterfreund,
  Svirsky, Casey, and Kluger]{lindenbaum2021differentiable}
Lindenbaum, O., Shaham, U., Peterfreund, E., Svirsky, J., Casey, N., and
  Kluger, Y.
\newblock Differentiable unsupervised feature selection based on a gated
  laplacian.
\newblock \emph{Advances in Neural Information Processing Systems},
  34:\penalty0 1530--1542, 2021{\natexlab{b}}.

\bibitem[Lundberg \& Lee(2017)Lundberg and Lee]{lundberg2017unified}
Lundberg, S.~M. and Lee, S.-I.
\newblock A unified approach to interpreting model predictions.
\newblock \emph{Advances in neural information processing systems}, 30, 2017.

\bibitem[Lv et~al.(2021)Lv, Kang, Lu, and Xu]{lv2021pseudo}
Lv, J., Kang, Z., Lu, X., and Xu, Z.
\newblock Pseudo-supervised deep subspace clustering.
\newblock \emph{IEEE Transactions on Image Processing}, 30:\penalty0
  5252--5263, 2021.

\bibitem[Nie et~al.(2010)Nie, Huang, Cai, and Ding]{nie2010efficient}
Nie, F., Huang, H., Cai, X., and Ding, C.
\newblock Efficient and robust feature selection via joint l2, 1-norms
  minimization.
\newblock \emph{Advances in neural information processing systems}, 23, 2010.

\bibitem[Niu et~al.(2021)Niu, Shan, and Wang]{niu2021spice}
Niu, C., Shan, H., and Wang, G.
\newblock Spice: Semantic pseudo-labeling for image clustering.
\newblock \emph{arXiv preprint arXiv:2103.09382}, 2021.

\bibitem[Peng et~al.(2022)Peng, Li, Tsang, Zhu, Lv, and Zhou]{peng2022xai}
Peng, X., Li, Y., Tsang, I.~W., Zhu, H., Lv, J., and Zhou, J.~T.
\newblock Xai beyond classification: Interpretable neural clustering.
\newblock \emph{The Journal of Machine Learning Research}, 23\penalty0
  (1):\penalty0 227--254, 2022.

\bibitem[Piloto \& Schilling(2010)Piloto and Schilling]{piloto2010ovo1}
Piloto, S. and Schilling, T.~F.
\newblock Ovo1 links wnt signaling with n-cadherin localization during neural
  crest migration.
\newblock \emph{Development}, 137\penalty0 (12):\penalty0 1981--1990, 2010.

\bibitem[Qian et~al.(2023)Qian, Cebere, and van~der Schaar]{qian2023synthcity}
Qian, Z., Cebere, B.-C., and van~der Schaar, M.
\newblock Synthcity: facilitating innovative use cases of synthetic data in
  different data modalities.
\newblock \emph{arXiv preprint arXiv:2301.07573}, 2023.

\bibitem[Reddy et~al.(2018)Reddy, Al~Hasan, and Zaki]{reddy2018clustering}
Reddy, C.~K., Al~Hasan, M., and Zaki, M.~J.
\newblock Clustering biological data.
\newblock \emph{Data Clustering}, pp.\  381--414, 2018.

\bibitem[Roe et~al.(2005)Roe, Yang, Zhu, Liu, Stancu, and
  McGregor]{roe2005boosted}
Roe, B.~P., Yang, H.-J., Zhu, J., Liu, Y., Stancu, I., and McGregor, G.
\newblock Boosted decision trees as an alternative to artificial neural
  networks for particle identification.
\newblock \emph{Nuclear Instruments and Methods in Physics Research Section A:
  Accelerators, Spectrometers, Detectors and Associated Equipment},
  543\penalty0 (2-3):\penalty0 577--584, 2005.

\bibitem[Sato et~al.(2009)Sato, Vries, Snippert, Van De~Wetering, Barker,
  Stange, Van~Es, Abo, Kujala, Peters, et~al.]{sato2009single}
Sato, T., Vries, R.~G., Snippert, H.~J., Van De~Wetering, M., Barker, N.,
  Stange, D.~E., Van~Es, J.~H., Abo, A., Kujala, P., Peters, P.~J., et~al.
\newblock Single lgr5 stem cells build crypt-villus structures in vitro without
  a mesenchymal niche.
\newblock \emph{Nature}, 459\penalty0 (7244):\penalty0 262--265, 2009.

\bibitem[Shaham et~al.(2018)Shaham, Stanton, Li, Nadler, Basri, and
  Kluger]{shaham2018spectralnet}
Shaham, U., Stanton, K., Li, H., Nadler, B., Basri, R., and Kluger, Y.
\newblock Spectralnet: Spectral clustering using deep neural networks.
\newblock \emph{arXiv preprint arXiv:1801.01587}, 2018.

\bibitem[Shaham et~al.(2022)Shaham, Lindenbaum, Svirsky, and
  Kluger]{shaham2022deep}
Shaham, U., Lindenbaum, O., Svirsky, J., and Kluger, Y.
\newblock Deep unsupervised feature selection by discarding nuisance and
  correlated features.
\newblock \emph{Neural Networks}, 152:\penalty0 34--43, 2022.

\bibitem[Shen et~al.(2021)Shen, Shen, Wang, Qin, Torr, and Shao]{shen2021you}
Shen, Y., Shen, Z., Wang, M., Qin, J., Torr, P., and Shao, L.
\newblock You never cluster alone.
\newblock \emph{Advances in Neural Information Processing Systems},
  34:\penalty0 27734--27746, 2021.

\bibitem[Singh et~al.(2002)Singh, Febbo, Ross, Jackson, Manola, Ladd, Tamayo,
  Renshaw, D'Amico, Richie, et~al.]{singh2002gene}
Singh, D., Febbo, P.~G., Ross, K., Jackson, D.~G., Manola, J., Ladd, C.,
  Tamayo, P., Renshaw, A.~A., D'Amico, A.~V., Richie, J.~P., et~al.
\newblock Gene expression correlates of clinical prostate cancer behavior.
\newblock \emph{Cancer cell}, 1\penalty0 (2):\penalty0 203--209, 2002.

\bibitem[Sokar et~al.(2022)Sokar, Atashgahi, Pechenizkiy, and
  Mocanu]{sokar2022pay}
Sokar, G., Atashgahi, Z., Pechenizkiy, M., and Mocanu, D.~C.
\newblock Where to pay attention in sparse training for feature selection?
\newblock \emph{arXiv preprint arXiv:2211.14627}, 2022.

\bibitem[Song et~al.(2013)Song, Liu, Huang, Wang, and Tan]{song2013auto}
Song, C., Liu, F., Huang, Y., Wang, L., and Tan, T.
\newblock Auto-encoder based data clustering.
\newblock In \emph{Progress in Pattern Recognition, Image Analysis, Computer
  Vision, and Applications: 18th Iberoamerican Congress, CIARP 2013, Havana,
  Cuba, November 20-23, 2013, Proceedings, Part I 18}, pp.\  117--124.
  Springer, 2013.

\bibitem[Sundararajan et~al.(2017)Sundararajan, Taly, and
  Yan]{sundararajan2017axiomatic}
Sundararajan, M., Taly, A., and Yan, Q.
\newblock Axiomatic attribution for deep networks.
\newblock In \emph{International conference on machine learning}, pp.\
  3319--3328. PMLR, 2017.

\bibitem[van Breukelen et~al.(1998)van Breukelen, Duin, Tax, and
  Den~Hartog]{van1998handwritten}
van Breukelen, M., Duin, R.~P., Tax, D.~M., and Den~Hartog, J.
\newblock Handwritten digit recognition by combined classifiers.
\newblock \emph{Kybernetika}, 34\penalty0 (4):\penalty0 381--386, 1998.

\bibitem[Varghese et~al.(2010)Varghese, Unnikrishnan, Sciencist, Kochi, and
  Kochi]{varghese2010clustering}
Varghese, B.~M., Unnikrishnan, A., Sciencist, G., Kochi, N., and Kochi, C.
\newblock Clustering student data to characterize performance patterns.
\newblock \emph{Int. J. Adv. Comput. Sci. Appl}, 2:\penalty0 138--140, 2010.

\bibitem[Vincent et~al.(2008)Vincent, Larochelle, Bengio, and
  Manzagol]{vincent2008extracting}
Vincent, P., Larochelle, H., Bengio, Y., and Manzagol, P.-A.
\newblock Extracting and composing robust features with denoising autoencoders.
\newblock In \emph{Proceedings of the 25th international conference on Machine
  learning}, pp.\  1096--1103, 2008.

\bibitem[Von~Luxburg(2007)]{von2007tutorial}
Von~Luxburg, U.
\newblock A tutorial on spectral clustering.
\newblock \emph{Statistics and computing}, 17:\penalty0 395--416, 2007.

\bibitem[Wang \& Bodovitz(2010)Wang and Bodovitz]{wang2010single}
Wang, D. and Bodovitz, S.
\newblock Single cell analysis: the new frontier in ‘omics’.
\newblock \emph{Trends in biotechnology}, 28\penalty0 (6):\penalty0 281--290,
  2010.

\bibitem[Xie et~al.(2016)Xie, Girshick, and Farhadi]{xie2016unsupervised}
Xie, J., Girshick, R., and Farhadi, A.
\newblock Unsupervised deep embedding for clustering analysis.
\newblock In \emph{International conference on machine learning}, pp.\
  478--487. PMLR, 2016.

\bibitem[Xu et~al.(2023)Xu, Wang, Nie, and Li]{xu2023efficient}
Xu, L., Wang, R., Nie, F., and Li, X.
\newblock Efficient top-k feature selection using coordinate descent method.
\newblock In \emph{Proceedings of the AAAI Conference on Artificial
  Intelligence}, volume~37, pp.\  10594--10601, 2023.

\bibitem[Yamada et~al.(2020)Yamada, Lindenbaum, Negahban, and
  Kluger]{yamada2020feature}
Yamada, Y., Lindenbaum, O., Negahban, S., and Kluger, Y.
\newblock Feature selection using stochastic gates.
\newblock In \emph{International Conference on Machine Learning}, pp.\
  10648--10659. PMLR, 2020.

\bibitem[Yang et~al.(2022)Yang, Lindenbaum, and Kluger]{yang2022locally}
Yang, J., Lindenbaum, O., and Kluger, Y.
\newblock Locally sparse neural networks for tabular biomedical data.
\newblock In \emph{International Conference on Machine Learning}, pp.\
  25123--25153. PMLR, 2022.

\bibitem[Yang et~al.(2021)Yang, Huang, and Liu]{yang2021feature}
Yang, P., Huang, H., and Liu, C.
\newblock Feature selection revisited in the single-cell era.
\newblock \emph{Genome Biology}, 22:\penalty0 1--17, 2021.

\bibitem[Yoon et~al.(2019)Yoon, Jordon, and van~der Schaar]{yoon2019invase}
Yoon, J., Jordon, J., and van~der Schaar, M.
\newblock Invase: Instance-wise variable selection using neural networks.
\newblock In \emph{International Conference on Learning Representations}, 2019.

\bibitem[Yu et~al.(2020)Yu, Chan, You, Song, and Ma]{yu2020learning}
Yu, Y., Chan, K. H.~R., You, C., Song, C., and Ma, Y.
\newblock Learning diverse and discriminative representations via the principle
  of maximal coding rate reduction.
\newblock \emph{Advances in Neural Information Processing Systems},
  33:\penalty0 9422--9434, 2020.

\bibitem[Zhao \& Liu(2012)Zhao and Liu]{zhao2012spectral}
Zhao, Z.~A. and Liu, H.
\newblock \emph{Spectral feature selection for data mining}.
\newblock Taylor \& Francis, 2012.

\bibitem[Zheng et~al.(2017)Zheng, Terry, Belgrader, Ryvkin, Bent, Wilson,
  Ziraldo, Wheeler, McDermott, Zhu, et~al.]{zheng2017massively}
Zheng, G.~X., Terry, J.~M., Belgrader, P., Ryvkin, P., Bent, Z.~W., Wilson, R.,
  Ziraldo, S.~B., Wheeler, T.~D., McDermott, G.~P., Zhu, J., et~al.
\newblock Massively parallel digital transcriptional profiling of single cells.
\newblock \emph{Nature communications}, 8\penalty0 (1):\penalty0 14049, 2017.

\end{thebibliography}
